\documentclass{article}

\usepackage{comment}
\usepackage{times}
\usepackage{microtype}
\usepackage{graphicx}
\usepackage{booktabs} 
\usepackage{multirow}
\usepackage{url}
\usepackage{pifont}
\usepackage{amsmath}
\usepackage{amssymb}
\usepackage{epsfig}
\usepackage{subcaption}
\usepackage{siunitx}

\usepackage[pagebackref=true,breaklinks=true,colorlinks,bookmarks=false]{hyperref}

\usepackage[accepted]{icml2021}

\icmltitlerunning{DeepReDuce:  ReLU Reduction for Fast Private Inference}

\begin{document}

\twocolumn[
\icmltitle{DeepReDuce:  ReLU Reduction for Fast Private Inference}




\begin{icmlauthorlist}
\icmlauthor{Nandan Kumar Jha}{aa}
\icmlauthor{Zahra Ghodsi}{aa}
\icmlauthor{Siddharth Garg}{aa}
\icmlauthor{Brandon Reagen}{aa}
\end{icmlauthorlist}

\icmlaffiliation{aa}{New York University, New York, USA}
%\icmlaffiliation{goo}{Googol ShallowMind, New London, Michigan, USA}
%\icmlaffiliation{ed}{School of Computation, University of Edenborrow, Edenborrow, United Kingdom}

\icmlcorrespondingauthor{Nandan Kumar Jha}{nj2049@nyu.edu}
%\icmlcorrespondingauthor{Eee Pppp}{ep@eden.co.uk}

\icmlkeywords{Machine Learning, ICML}

\vskip 0.3in
]



\printAffiliationsAndNotice{}  % leave blank if no need to mention equal contribution

\begin{abstract}

The recent rise of privacy concerns has led researchers to devise
methods for private neural inference---where 
inferences are made directly on encrypted data, never seeing inputs.
The primary challenge facing private inference is that computing on encrypted data
levies an impractically-high latency penalty, stemming mostly from non-linear operators like ReLU.
Enabling practical and private inference requires new optimization methods that
minimize network ReLU counts while preserving accuracy. 
This paper proposes \textit{DeepReDuce}: a set of optimizations for
the judicious removal of ReLUs to reduce private inference latency.
The key insight is that not all ReLUs contribute equally to accuracy.
We leverage this insight to drop, or remove, ReLUs
from classic networks to significantly 
reduce inference latency and maintain high accuracy.
Given a network architecture,
DeepReDuce outputs a Pareto frontier of networks
that tradeoff the number of ReLUs and accuracy.
Compared to the state-of-the-art for private inference
DeepReDuce improves accuracy and reduces ReLU count by up to
3.5\% (iso-ReLU count) and 3.5$\times$ (iso-accuracy), respectively.

\end{abstract}

\section{Introduction}

%\fixme{Decide the terminology: ReLU pruning or ReLU dropping}
% Privacy is important
Concerns surrounding data privacy continue to rise 
and are beginning to affect technology.
Companies are changing the way they use and store users' data
while lawmakers are passing legislation to improve users' privacy rights~\cite{hipaa, gdpr}.
Deep learning is the core driver of many applications impacted by privacy concerns.
It provides high utility in classifying, recommending, and interpreting user
data to build user experiences and requires large 
amounts of private user data to do so.
Private inference is a solution that simultaneously 
provides strong privacy guarantees while preserving the utility of neural networks 
to power application experiences users enjoy.
Today, in a typical inference pipeline, 
a client encrypts data (ciphertexts) and sends it to the cloud,
the cloud decrypts the data and performs inferences on the data (plaintext),
and returns the encrypted results.
With private inference, the same inference is executed
without the cloud ever decrypting the client's data;
that is, inferences are processed directly on ciphertexts.

\begin{figure}[t] \centering
\includegraphics[scale=0.45]{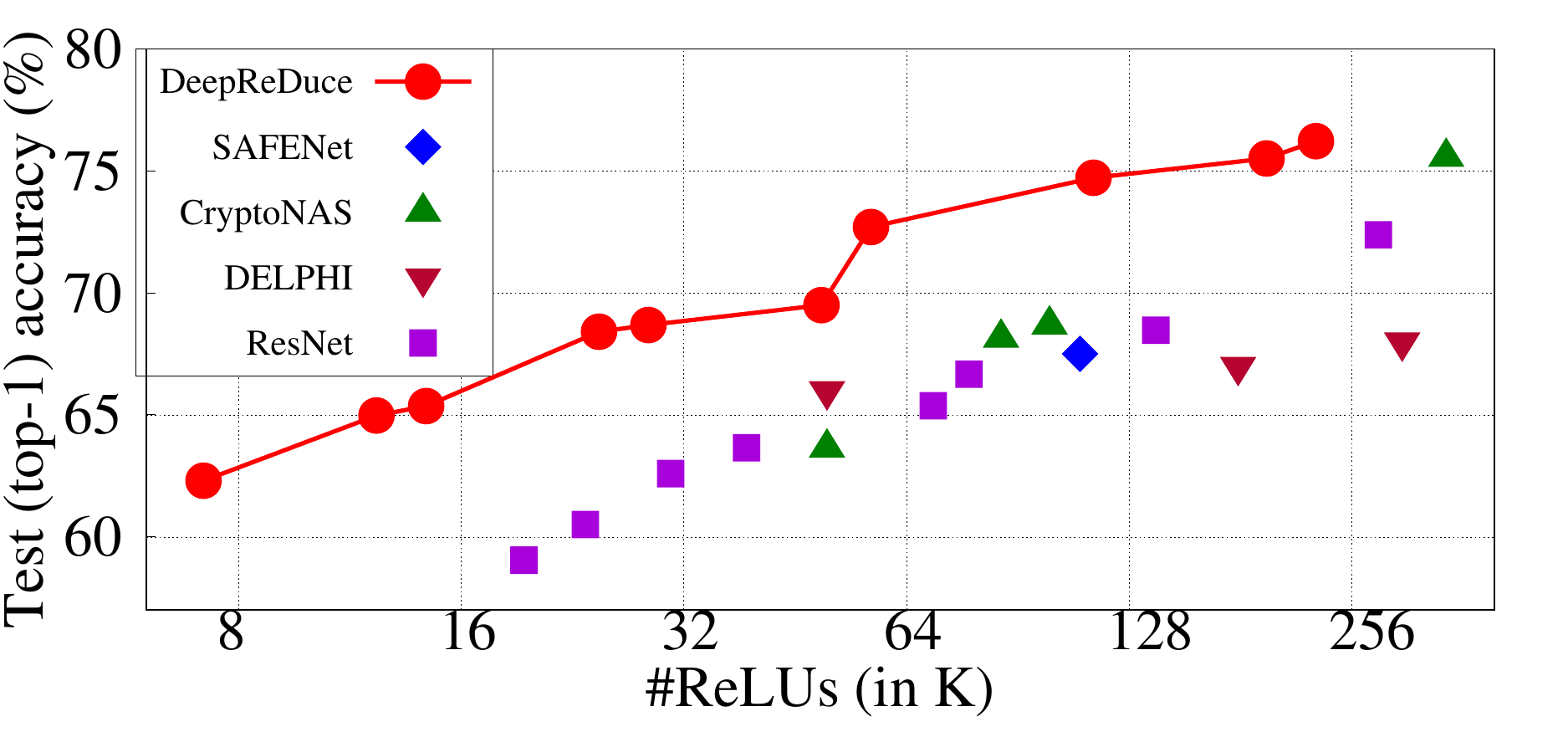}
\vspace{-2em}
\caption{DeepReDuce Pareto frontier of ReLU counts versus accuracy for CIFAR-100.
We show DeepReDuce outperforms
the state-of-the-art (SAFENet \cite{lou2021safenet}, CryptoNAS \cite{ghodsi2020cryptonas} and DELPHI \cite{mishra2020delphi}).}
%for private inference.}
\vspace{-2em}
\label{fig:ParetoFrontier}
\end{figure}

% Approaches and limitations
Prior work has established two methods for private inference.
The primary difference between them is how linear layers are computed,
either with secret-sharing (SS), a type of secure multi-party computation (MPC) \cite{goldreich2019play,shamir1979share}, or homomorphic encryption (HE), an encryption scheme that enables computation on ciphertexts \cite{gentry2009fully,brakerski2014efficient}.
For example, Gazelle~\cite{juvekar2018gazelle} and Cheetah~\cite{br2020cheetah} use HE for convolution and fully-connected layers
whereas MiniONN~\cite{liu2017oblivious}, DELPHI~\cite{mishra2020delphi}, and CryptoNAS~\cite{ghodsi2020cryptonas}
use SS. 
While distinct, both offer limited functional support and cannot readily process non-linear operations, e.g., ReLUs. 
To overcome this limitation, private inference protocols use Garbled circuits (GCs), another form of MPC \cite{yao1982protocols,yao1986generate}, to process ReLU privately.
%Switching between HE or SS to
However, GCs are several orders of magnitude more expensive than the linear layer protocols in terms of communication and computational, rendering private inference impractical~\cite{ghodsi2020cryptonas}.
For example, ReLUs account for 93\% of ResNet32's online private inference time using the DELPHI protocol~\cite{mishra2020delphi}.
Therefore, enabling low-latency private inference is a matter of minimizing a network's ReLU count. 

% Core contribution
%This paper investigates the relationship between the 
%ReLU count of a neural network and its overall accuracy.

To optimize networks for ReLU count we propose \emph{ReLU dropping}.
Inspired by weight and channel pruning methods that improve plaintext inference speed by reducing FLOPs, 
ReLU dropping directly reduces ReLU counts by removing entire ReLU layers from the network.
While weight and channel pruning also reduce ReLU counts, their ReLU savings are modest compared to FLOP savings.
%\footnote{For instance, scaling the number of channels in each layer by $\alpha$ reduces ReLU counts by $\alpha$ but FLOPs by $\alpha^{2}$.}.
In contrast, ReLU dropping achieves large reductions in ReLU counts by 
removing ReLU layers wholesale.
Leveraging our observations that
ReLU operators are unevenly distributed across network layers and
contribute differently to accuracy, we find ReLU dropping can be effectively 
applied to networks for large ReLU count reductions with minimal impact on accuracy.

%because even linear reductions in ReLU counts
%ReLU dropping selectively removes ReLUs operators to speedup private inference.
%While pruning methods also reduce ReLU counts, they only achieve modest reductions in 
%primary focus is on FLOPs; 
%cutting the number of channels in each layer by $2\times$ reduces ReLU count by $2\times$ but FLOPs by $4\times$
%Prior work has focused on replacing ReLUs with SS/HE amenable activations
%(e.g., X$^{2}$~\cite{gilad2016cryptonets,mishra2020delphi}) or proposed new network architectures that maximize weights per ReLU~\cite{ghodsi2020cryptonas}.
%This paper takes a novel and more direct approach by removing ReLUs wholesale.
%based on their 
%criticality, i.e., impact on the accuracy of network.
%ReLU dropping builds on our observations, described in Section~\ref{sec:Method}, that
%ReLU operators are unevenly distributed across network layers,
%contribute differently to accuracy,
%and that large fractions can be removed
%without significant accuracy loss.
%We show that 
%\fixme{SG: following sentence repetitive. remove?}
%Because ReLU operators dominate private inference run-time~\cite{ghodsi2020cryptonas},
%the optimizations result in significant latency improvement.

%We rigorously evaluate DeepReDuce using CIFAR-100, 
%the standard benchmark used for private inference~\cite{mishra2020delphi, ghodsi2020cryptonas}.
%Figure~\ref{fig:ParetoFrontier} shows that DeepReDuce's networks dominate 
%the ReLU count-accuracy Pareto frontier compared to CryptoNAS and DELPHI,  
%the current state of the art in private inference.

To further improve results we synergistically combine ReLU dropping with knowledge distillation (KD) \cite{hinton2015distilling,wang2021knowledge} to maximize the accuracy of optimized networks using the original Full-ReLU network.
We refer to our overall methodology as \emph{DeepReDuce}, 
which includes both optimizations for ReLU dropping and KD training.
%a method to optimize networks for private inference by selectively removing entire ReLU \fixme{layers}.
Figure~\ref{fig:ParetoFrontier} compares DeepReDuce against the current state-of-the-art: 
SAFENet \cite{lou2021safenet}, a method for selectively replacing channel-wise ReLUs with multiple degree and layer-wise mixed precision polynomials;
CryptoNAS~\cite{ghodsi2020cryptonas}, a neural architecture search (NAS) method targeting private inference and 
DELPHI~\cite{mishra2020delphi} a method for selectively substituting layer-wise ReLU activations with degree two polynomial activation functions.
We find that DeepReDuce significantly advances the accuracy-ReLU budget Pareto frontier across a wide design space.

% Discussion, answer obvious questions: Cant just train smaller network?
ReLU dropping has the added benefit that
when ReLUs are removed, the now adjacent linear transformations can be combined or merged, 
reducing the model depth and overall computations, i.e., FLOPs.
One obvious alternative to ReLU dropping is to simply start with shallower networks.
Our experiments show that DeepReDuce outperforms training shallower networks.
For example, ResNet9 (a down-scaled version of ResNet18, details in Section~\ref{sec:ExperimentalResults}) uses 30,720 ReLUs and and achieves 66.2\% accuracy 
whereas DeepReDuce produces a network with both \emph{fewer} ReLUs (24,600 ReLUs) and \emph{higher} accuracy (68.1\%).
%We note that the granularity of ReLU dropping is coarse.
%Therefore, we report numbers that show benefit in both ReLU count and accuracy.
We believe this is due to the fact that large models train better, 
which was also observed in \cite{zhao2018reThinking}. Detailed discussion is included in Section~\ref{subsec:ShallowResNet}.

% With respect to weight pruning, we first note that irregular weight pruning, such as~\cite{}, will not help as the distribution of remaining non-zero weights are random making it difficult to remove a significant number of ReLUs.
% Recent proposals for channel pruning~\cite{}, however, could help.
% In fact, one could interpret ReLU dropping as an extreme instance of channel pruning, 
% i.e., when all channels are pruned the effect is the same as removing a layer of ReLUs.
% However, channel pruning is not competitive with DeepReDuce and takes substantially longer to run.
% \fixme{We need data for this and to strengthen the text.
% Nandan, please enter your numbers.} \fixme{SG: agreed, this is critical imo b/c its the first in a reviewer would have.}

This paper makes the following contributions:
\begin{enumerate}
    \item Motivate the proposed idea of ReLU dropping and develop DeepReDuce:
            a method for the judicious removal of ReLUs to optimize networks
            for fast and accurate private inference.

    \item Rigorous evaluation of DeepReDuce demonstrating Pareto optimal designs across a wide range of       accuracy and ReLU counts. 
        DeepReDuce improves accuracy up to     
        3.5\% (iso-ReLU count) and reduces ReLUs by 3.5$\times$ (iso-accuracy) over the state of the art.

    %\item We conclude with a case study using TinyImageNet to demonstrate the generality of ReLU   
    %dropping beyond CIFAR-100 and ResNet.

    \item Show existing techniques for neural inference efficiency are insufficient for ReLU reduction and private inference. E.g., compared to the state-of-the-art channel pruning technique~\cite{he2020learning}, DeepReDuce provides a 2$\times$ greater ReLU reduction at with similar accuracy.
    %and provide a discussion of trade-offs in designing networks for small ReLU budgets.
    \end{enumerate}

%\fixme{Shall we add the performance comparison with SOTA pruning method here in contribution?}

\section{Motivating ReLU Dropping} \label{sec:Motivation}

In this section we motivate and present the key intuition behind ReLU dropping.  
We begin by defining the terms relevant to our discussion.

\begin{figure}[htbp] \centering
\includegraphics[scale=0.55]{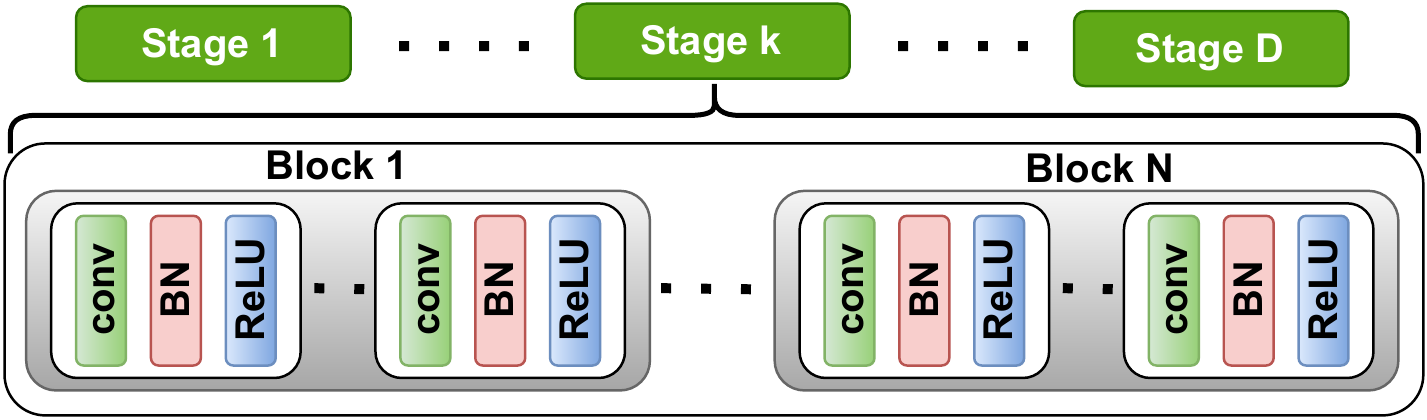}
\vspace{-1em}
\caption{The structure of conventional ResNet-like architectures: %\cite{he2016deep,zagoruyko2016wide,
%xie2017aggregated,huang2017densely,huang2018condensenet,brendel2018approximating}.
Stages comprise Blocks and Blocks comprise multiple repetitions of conv, BN, and ReLU layers. 
%We follow this hierarchy in DeepReDuce in order to reduce the huge search space for ReLU optimization.
%this is the terminology we use to describe how ReLU dropping optimizations are applied.
}
\vspace{-1em}
\label{fig:NetHierarchy}
\end{figure}

\subsection{Notation}
Many state-of-art DNNs have a well defined hierarchy, 
which allows them to easily scale to different design points \cite{he2016deep,zagoruyko2016wide,xie2017aggregated,huang2017densely,huang2018condensenet,brendel2018approximating,sandler2018mobilenetv2}. 
These architectures consist of multiple Stages ($S$)
and each Stage contains copies of the same Block ($B$), as shown in Figure \ref{fig:NetHierarchy}. 
It is typical to call out the first convolution layer as Conv1, which we also do here.
The spatial resolution of the feature maps (fmaps) is same within a Stage. 
For example, the ResNet18 architecture contains four Stages, 
each with two residual Blocks where residual Blocks constitute
two $3\times3$ convolution layers \cite{he2016deep}.
Conventional scaling methods for designing smaller networks include 
channel and feature map scaling.
Channel scaling reduces the dimensions of the weights by a factor $\alpha$
and feature map scaling reduces the input resolution by $\rho$~\cite{howard2017mobilenets,tan2019efficientnet}.

When describing a DeepReDuce optimized network we explicitly name stages with ReLUs intact.
E.g., $S_{2} + S_{3}$ implies stages $S_1$ and $S_4$ have their ReLUs completely removed and
only $S_2$ and $S_3$ have ReLUs.
When a stage is optimized with \textit{ReLU Thinning} (see below for details)
we superscript it with $RT$, e.g., $S_2^{RT}$.
When channel and feature maps' resolution scaling are applied we specify $\rho$ and $\alpha$ amounts.
%\fixme{This is not consistent with algorithm1.}
 \begin{table} [t]
\caption{ReLUs' criticality evaluation: ReLU counts and accuracy (CIFAR-100) for ResNet models
where ReLUs are dropped from all but one stage. 
%E.g., S1 indicates the only network ReLUs are in S1.
We posit that less accurate ReLU stages indicate less important
ReLUs and note accuracy differs significantly across stages.}
\label{tab:ReluHeteroR18}
\centering 
\resizebox{0.49\textwidth}{!}{
\begin{tabular}{cccccccc} \toprule
Models & Metrics & No ReLUs & Conv1 & $S_1$ & $S_2$ & $S_3$ & $S_4$ \\ \toprule
\multirow{3}{*}{ ResNet18 } & \#ReLUs & 0 & 66K & 262K & 131K & 66K & 33K \\
& W/o KD (\%) & 18.49 & 46.22 & 61.93 & 67.63 & 67.41 & 58.90 \\
& W/ KD (\%) & 18.34 & 45.07 & 59.85 & 68.79 & 69.92 & 63.16 \\ \midrule
\multirow{3}{*}{ResNet34} & \#ReLUs & 0 & 66K & 393K & 262K & 197K & 49K \\
& W/o KD(\%) & 18.16 & 45.42 & 60.77 & 69.47 & 70.04 & 57.44 \\
& W/ KD(\%) & 18.07 & 45.13 & 62.88 & 70.93 & 72.61 & 64.23 \\
\bottomrule
\end{tabular} }
\end{table}

\subsection{ReLU Dropping}
We now discuss four observations motivate ReLU dropping and DeepReDuce.

\begin{figure}[t]
\includegraphics[scale=0.19]{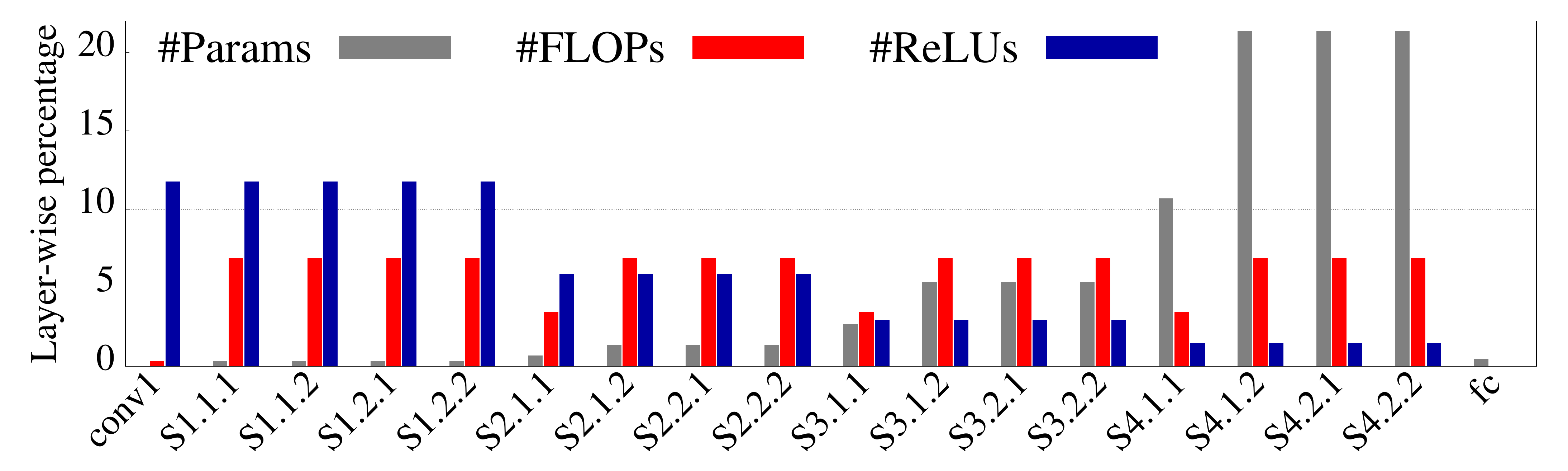}
\vspace{-2em}
\caption{
Layer-wise distribution of parameters, FLOPs, and ReLUs in ResNet18.
FLOPs are evenly distributed, parameters (ReLUs) are increases (decreases) with network's depth.} 
\label{fig:LayerWiseReluInDNNs}
\vspace{-1em}
\end{figure}

\textbf{Observation 1: ReLUs are unevenly distributed across
the layers of conventional CNNs.} 
We begin by investigating the distribution of ReLUs across the layers of modern CNN architectures. 
Figure~\ref{fig:LayerWiseReluInDNNs} presents a layer-wise breakdown
of ResNet18's ReLUs, FLOPS, and parameters.
We observe that FLOPs are evenly distributed across layers,
and that the number of ReLUs per layer decreases with depth while 
the parameter count increases with depth.
A similar distribution of parameters, FLOPs, and ReLUs have
been observed in other common CNNs
(Figure \ref{fig:LayerWiseReluInOtherDNNs} in  Appendix \ref{secAppendix:LayerWiseOpsInDNNs}).

The skewed distribution of ReLUs is because
conventional CNN architectures
tend to scale the number of channels up by $2\times$ and the fmap spatial dimension down by $2\times$ in \emph{each} dimension across stages, 
resulting in a $2\times$ drop in ReLU count across a down-sampling layer.
All other Things being equal, this presents an opportunity to significantly reduce
ReLU counts by simply dropping them from early stages.
This fortuitous as we also observe
ReLUs in the early stages tend to be less critical for accuracy.

\textbf{Observation 2: ReLUs in some stages are more 
important for accuracy than others.}
To understand the relative importance of ReLUs in different network stages
we crafted ablation experiments.
This was done by removing ReLUs from all but one ResNet18 stage
and training each resulting network from scratch.
Table~\ref{tab:ReluHeteroR18} shows the accuracy and ReLU count of each resulting network
using the CIFAR-100 dataset. 
Recall that ResNet18 has Conv1 layer prior to stage 1.
For completeness (in Table~\ref{tab:ReluHeteroR18}) we report result for ResNet18 with ReLUs only in Conv1.  However, we always drop ReLUs from Conv1 along with stages in the network in subsequent experiments.

We note that the four resulting networks vary greatly with respect to accuracy. 
Networks with ReLUs in $S_2$ and $S_3$ (Table \ref{tab:ReluHeteroR18}) have high accuracy, 
even though $S_2$ and $S_3$ use fewer ReLUs than $S_1$. 
Similarly, although Conv1 and $S_3$ have the same number of ReLUs, 
allocating ReLUs to $S_3$ instead of Conv1 increases accuracy by 24.8\%. 
A similar ReLU-accuracy disparity was observed for ResNet34 (see Table~\ref{tab:ReluHeteroR18}.) 
We conclude that not all ReLUs are equal:
some contribute more to model accuracy than others.
We hypothesize these less important ReLUs can be removed without significantly impacting network's accuracy.

{\bf Observation 3: Some ReLUs benefit more from knowledge distillation that others.}
Given the disparate impact of dropping ReLUs from stages in a network, 
we also investigated whether some stages benefit more from KD than others. 
As illustrated in the Table~\ref{tab:ReluHeteroR18}, 
the accuracy gain from KD is position dependent and greater for networks with ReLUs in deeper stages ($S_4$ and $S_3$) compared to the networks with ReLUs in initial stages ($S_1$ and $S_2$).
Our results thus suggest that KD is synergistic with ReLU dropping:
dropping ReLUs from early layers dramatically reduces ReLU count with a relatively small impact on accuracy and would not have benefited as much from KD had ReLUs in these layers been preserved.
Conversely, latter layers with a small number of more critical ReLUs also benefit the most from KD. 
This resonates with the claim made in~\cite{gotmare2018a}
as knowledge shared by a teacher in KD is primarily disbursed in deeper layers.
 
%are the easiest to drop, have the most ReLUs, and are ill-suited for 
%KD while later layers have fewer ReLUs and work well with KD.

\begin{table} [t]
\caption{
Performance comparison of channel scaling ($\alpha$=0.5) and dropping ReLUs 
from alternate layers (S$_k^{RT}$ where $k$ is the network stage) using ResNet18 on CIFAR-100. 
Both methods reduce ReLUs by a factor of 2$\times$; 
however, alternate ReLU dropping results in more accurate iso-ReLU networks.}
\label{tab:AlphaVsReluDrop}
\centering 
\resizebox{0.49\textwidth}{!}{
\begin{tabular}{ccccc} 
 \toprule
 Network & \#Conv & \#ReLUs & W/o KD(\%) & W/ KD(\%) \\ \toprule
%\multirow{1}{*}{\bf Network } & \multirow{1}{*}{\bf \#Conv } & \multirow{1}{*}{\bf \#ReLU } & \multicolumn{2}{c}{\bf CIFAR-100 } \\
%& & & {\bf w/o kD} & {\bf w/ KD} \\
 
$S_{2}$+$S_{3}$+$S_{4}$  & 17 & 229K & 73.14 & 76.22 \\
$S_{2}^{RT}$+$S_{3}^{RT}$+$S_{4}^{RT}$ & 17 & 115K & 72.97 & 74.72 \\
$S_{2}$+$S_{3}$+$S_{4}$, $\alpha$=0.5 & 17 & 115K & 71.59 & 73.78 \\ \midrule
%$Arch2$ $\rightarrow$ Res18(FR) - relu[conv1 + S1] & 11 & 115K & 71.96 &	74.57 \\ 
$S_{2}$+$S_{3}$  & 17 & 197K & 72.77 & 75.51 \\
$S_{2}^{RT}$+$S_{3}^{RT}$ & 17 & 98K & 70.97 & 71.95 \\
$S_{2}$+$S_{3}$, $\alpha$=0.5 & 17 & 98K & 69.54 & 71.16 \\ \midrule
%$Arch2$ $\rightarrow$ Res18(FR) - relu[conv1 + S1 +S4] & 13 & 98K & 70.15 &	72.17 \\ \midrule
$S_{3}$+$S_{4}$ & 17 & 98K & 68.4 & 73.16 \\
$S_{3}^{RT}$+$S_{4}^{RT}$ & 17 &49K & 69.62 & 71.06 \\
$S_{3}$+$S_{4}$, $\alpha$=0.5 & 17 & 49K & 66.43 & 70.29 \\ 
%$Arch2$ $\rightarrow$ Res18(FR) - relu[conv1 + S1 + S2] & 13 & 49K & 66.68 &	69.79 \\
\bottomrule
\end{tabular}} 
\vspace{-1.5em}
\end{table}

\textbf{Observation 4: Dropping ReLUs within stages is better than channel scaling.}
Our final observation relates to the relative importance of ReLUs \emph{within} network 
stages. 
We explore two strategies for $2\times$ ReLU reduction in a stage: 
(i) drop ReLUs from alternate layers or 
(ii) reduce the number of channels in the stage by $2\times$.
Table~\ref{tab:AlphaVsReluDrop} compares alternate layer ReLU dropping against channel down scaling for three different network architectures obtained by dropping ReLUs in $S_1$, $S_1+S_4$, and $S_1+S_2$.
In each instance we observe that at iso-ReLU, alternate layer ReLU dropping has $1\%-3\%$ accuracy improvements over channel down-scaling. 
We note that even with KD
alternate layer ReLU dropping is consistently better than channel down-scaling.
\begin{figure*}[htbp] \centering
\includegraphics[scale=0.55]{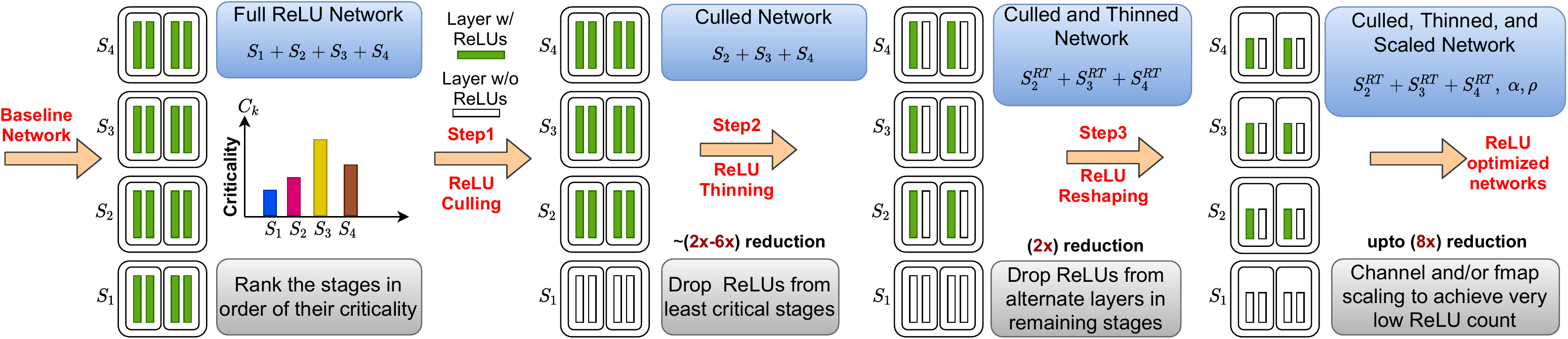}
\caption{The DeepReDuce ReLU optimization pipeline.
The baseline is optimized left to right following orange arrows.
Green (plain) boxes indicate layer's with (without) ReLUs, gray boxes indicate profiling steps to guide optimizations, and blue boxes describe the resulting network after each optimization step. Given a baseline network, DeepReDuce outputs ReLU-optimized networks preserving as much accuracy from the baseline network
as possible.}
\label{fig:BlockDiagramDeepReDuce}
\end{figure*}

\section{DeepReDuce} 
\label{sec:Method}

Using the insights listed above, this section presents the proposed ReLU optimizations and
resulting method for optimizing networks for private inference.
Given a baseline network as input,
DeepReDuce outputs a set of Pareto optimal networks 
that trade accuracy and ReLU count.
This is done via three ReLU reducing optimizations 
(shown as Step 1 to 3 in Figure \ref{fig:BlockDiagramDeepReDuce}):
Culling (Section \ref{sec:culling}), Thinning  (Section~\ref{sec:Thinning}), and Reshaping (Section~\ref{sec:Reshaping}). 
The optimizations work at different levels of granularity and are applied sequentially
to identify high-performing networks quickly.

\subsection{ReLU Culling}
\label{sec:culling}

The first optimization, named \textit{ReLU Culling}, 
removes ReLUs at a coarse granularity (Step 1 in Figure \ref{fig:BlockDiagramDeepReDuce}).
It works by completely removing all ReLUs from a given network stage.
Culling is applied iteratively,
from least to most critical using a criticality metric (described below)
that estimates the importance of each stage's ReLUs.  
Empirically we find the initial stage tends to be highly amenable to Culling as it 
contains a large number of non-critical ReLUs.
For a network with $D$ stages, ReLU Culling outputs $D-1$ networks that trade accuracy for reduced ReLU counts.

{\bf Criticality metric:} 
The order that a network's stages are culled is determined by estimating ReLU criticality,
i.e., how important a particular stage's ReLUs are for accuracy.
We denote each stage's ($S_k$) criticality as $C_k$.
To determine criticality, we first remove all the ReLUs from all network stages except $S_k$,
train the network, measure its accuracy, and then repeat the process with KD using the original network as a teacher. 
The criticality metric $C_k$ captures the accuracy improvement from retaining ReLUs in stage $S_k$ and the stage's ReLU cost as follows:

\begin{equation} \label{eqn:CriticalityMetric}
C_k = \frac{Acc [S_k] - \min_{(i=1 \; to \; D)} \{ Acc[Si] \}}{(\#ReLU [S_k])^{w}}  
\end{equation}

In Equation~\ref{eqn:CriticalityMetric}, $Acc [S_k]$ is the accuracy of the network with ReLUs only in stage $S_k$ trained with KD. 
$\#ReLU [S_k]$ in the denominator is the number of ReLUs in stage $S_k$. 
The hyper-parameter $w$ controls the weighted importance of accuracy and ReLU count, 
we set $w$=0.07, similar to \cite{tan2019efficientnet}.
The value of $C_k$ corresponding to least critical stage would be zero (see Table \ref{tab:ResNetBlockDropout}) and stages with higher $C_k$ utilize ReLUs better than stages with lower $C_k$. Hence, they are more critical and dropped later in DeepReDuce. We note that, even for the same network, criticality of ReLUs varies across different datasets. For instance, as shown in Table \ref{tab:ResNetBlockDropout} the criticality order of ResNet18 on CIFAR-100, from least to most critical, is $S_1 < S_4 < S_2 < S_3$ (calculated from the Table \ref{tab:ReluHeteroR18}); whereas, the same on TinyImageNet is $S_1 < S_2 < S_4 < S_3$.

\subsection{ReLU Thinning}
\label{sec:Thinning}
To further reduce ReLU counts, each Culled network from the ReLU Culling stage is further Thinned 
by dropping alternate ReLU layers in the remaining non-Culled stages of the network 
(Step 2 in Figure \ref{fig:BlockDiagramDeepReDuce}). This yields an additional reduction in ReLU counts for each Culled networks as the number of ReLU layers are halved.
%reduced by $2\times$. 

%\textcolor{red}{when layers in non-Culled Stages are symmetrical and having equal \#ReLUs (e.g., ResNet18). However, when alternate layers are asymmetrical and have unequal \#ReLUs, the reduction in ReLU count depends on the layers which have following ReLUs (e.g., MobileNetV1 \cite{howard2017mobilenets}). The detailed discussion is included in the Appendix \ref{SecAppendix:ParetoPointsMV1}.} 
  
Note that the same reduction in ReLU counts can be achieved via other means, for example, 
by dropping ReLUs from the first half or last half of a stage, or  
by scaling the number of channels in each 
layer by $2\times$. 
However, we empirically find that ReLU Thinning consistently outperforms both 
approaches (see results in Section~\ref{sec:Motivation}). 
Moreover, our empirical findings resonate with the observations made in~\cite{zhao2018reThinking}, which claims that removing ReLUs from alternate layers regularize the network and prevent information loss.

While Thinning could be generalized to drop ReLUs from arbitrary layers in non-Culled stages, this would significantly increase search complexity. Our choice of dropping ReLUs from alternate layers in non-Culled stages is to strike a balance between run-time and effectiveness of DeepReDuce.

\begin{table} [t] \centering
\caption{ReLUs' Criticality on TinyImageNet for ResNet18/34. FR is baseline with Full-ReLU ($S_1$+$S_2$+$S_3$+$S_4$). Similar to the our observations on CIFAR-100 in Table \ref{tab:ReluHeteroR18}, accuracy differs significantly across stages and $S_1$ ($S_3$) ReLUs are least (most) critical.}
\label{tab:ResNetBlockDropout} 
\resizebox{.49\textwidth}{!}{
\begin{tabular}{ccccccccc} \toprule
\multirow{2}{*}{ Net } & \multicolumn{4}{c}{ ResNet18 } & \multicolumn{4}{c}{ ResNet34 } \\ 
\cmidrule(lr{0.5em}){2-5}  
\cmidrule(lr{0.5em}){6-9} 
& \#ReLUs & W/o KD(\%) & W/ KD(\%) & $C_k$ &\#ReLUs & W/o KD(\%) & W/ KD(\%) & $C_k$ \\ \toprule
FR & 2228K & 61.28 & - &    -   &3867K & 63.06 & -     &  - \\ 
$S_1$ & 1049K & 41.90 & 39.61 &  0.00  &1573K & 42.10 & 39.4 & 0.00\\
$S_2$ & 524K & 50.53 & 49.44 & 6.04   &1049K & 53.49 & 51.74 & 7.58\\
$S_3$ & 262K & 51.93 & 54.34 &  9.50  &786K & 57.28 & 60.83 & 13.44 \\
$S_4$ & 131K & 46.89 & 51.46 &  8.02  &197K & 48.10 & 54.41 & 10.37 \\ \bottomrule
\end{tabular} }
\end{table}

% & stages & \#ReLUs & w/o KD & w/ KD \\ \toprule
%\multirow{9}{*}{ \rotatebox[origin=c]{90}{ResNet18} } &$S_1$ + $S_2$ + $S_3$ + $S_4$ & 2228K & 61.28 & - \\ \cline{2-5}
%& $S_1$ + $S_2$ + $S_3$& 1836K & 58.92 & 60.78 \\
%& $S_1$ + $S_2$ + $S_4$ & 1704K & 59.42 & 62.98 \\
%& $S_1$ + $S_3$ + $S_3$ & 1442K & 60.12 & 64.45 \\
%& $S_2$ + $S_3$ + $S_4$ & 918K & 60.5 & 64.66 \\  \cline{2-5}
%& $S_1$ & 1049K & 41.9 & 39.61 \\
%& $S_2$ & 524K & 50.53 & 49.44 \\
%& $S_3$ & 262K & 51.93 & 54.34 \\
%& $S_4$ & 131K & 46.89 & 51.46 \\ \midrule
%\multirow{9}{*}{ \rotatebox[origin=c]{90}{ResNet34} } & $S_1$ + $S_2$ + $S_3$ + $S_4$ & 3867K & 63.06 & - \\  \cline{2-5}
%& $S_1$ + $S_2$ + $S_3$ & 3408K & 61.93 & 65.01 \\
%& $S_1$ + $S_2$ + $S_4$ & 2818K & 61.19 & 64.73 \\
%& $S_1$ + $S_3$ + $S_3$ & 2556K & 62.2 & 65.92 \\
%& $S_2$ + $S_3$ + $S_4$ & 2032K & 63.45 & 66.69 \\  \cline{2-5}
%& $S_1$ & 1573K & 42.1 & 39.4 \\
%& $S_2$ & 1049K & 53.49 & 51.74 \\
%& $S_3$ & 786K & 57.28 & 60.83 \\
%& $S_4$ & 197K & 48.1 & 54.41 \\

\subsection{ReLU Reshaping}
\label{sec:Reshaping}
The final optimization of DeepReDuce uses conventional channel and fmaps resolution scaling
to decrease ReLU counts by reducing network size and shape 
(Step 3 in Figure \ref{fig:BlockDiagramDeepReDuce}).
While less effective than Culling and Thinning, as they tend to introduce higher accuracy drop, 
these optimization are useful in producing networks for highly-constrained ReLU budgets. 

To down-scale networks we explore three alternatives: 
channel scaling, feature map (fmap) scaling, and compound scaling.
Channel and fmap-resolution scaling reduce the filter count and spatial dimensions of fmaps across the network by factors of $\alpha$ and $\rho$, respectively. 
Compound scaling scales both channels and fmap spatial dimensions simultaneously, achieving multiplicative reductions in ReLU count. 
More precisely, channel and fmap resolution scaling by $\alpha$ and $\rho$ reduce the ReLU count by   $\alpha$ and $\rho^2$ respectively.
Since our aim is to gradually reduce the ReLU count, we first employ channel scaling ($\alpha$=0.5) and then fmap scaling ($\rho$=0.5) followed by compound scaling ($\alpha$=0.5 and $\rho$=0.5) to reduce the ReLU count by 2$\times$, 4$\times$, and 8$\times$, respectively. 

One can use different scaling factors for different degree of ReLU reduction.
However, naively selected scaling factors could result in suboptimal ReLU networks. 
For instance, $\alpha$=0.25 and $\rho$=0.5 both lower the ReLU count by 4$\times$; however, the former produces less accurate networks (see accuracy with KD in Table \ref{tab:AlphaVsRhoComp} in Appendix \ref{secAppendix:ChannelVsFmapScaling}). One possible explanation for the lower accuracy in channel-scaled networks can be the lower parameter count. 
That is, while $\alpha$=0.25 and $\rho$=0.5 reduces the ReLU count by same degree the former also reduces the parameter count by 4$\times$, which can reduce the expressiveness of the network.

We note that because Reshaping is
applied to all stages equally, it scales down the sizes of critical layers as well. 
As such, applying Reshaping earlier would reduce opportunities for our more effective Culling and Thinning optimizations. 
This is why we use Reshaping only as a last resort to reduce ReLU count.

\subsection{Improving accuracy using KD}
To maximize the accuracy of Culled, Thinned, and Reshaped networks we employ
knowledge distillation (KD) as the final step of DeeReDuce. 
Specifically, we re-train ReLU-optimized networks with Full-ReLU baseline  as a teacher, and
find distillation typically recovers several percentage points of accuracy on our datasets.
We note that although KD is only explicitly used as a final step, it is implicitly incorporated in the evaluation of stage criticality (see Section~\ref{sec:culling}), and guides the selection of which stages to Cull first (or last). 
Since the gain in accuracy from KD depends on the position of stages (Table \ref{tab:ReluHeteroR18} and \ref{tab:ResNetBlockDropout}), order of stage criticality computed with KD is different compared to that computed without KD. 
Therefore, incorporating KD ($Acc [S_k]$ in Eq. \ref{eqn:CriticalityMetric}) in computing criticality produces better results.

\subsection{Putting it All Together}

We developed DeepReDuce to effectively apply the above optimizations without exhaustively exploring the design space.
Given a network as input, DeepReDuce first determines the criticality of each stage.
We use this information to guide the application of coarse-grained ReLU Culling.
DeepReDuce iteratively applies Culling to each network stage from least to most critical.
The application of Culling is compounded after each iteration,
e.g., if S$_{2}$ was Culled first and S$_{3}$ is the next least critical stage, 
then DeepReDuce Culls both S$_{2}$ and S$_{3}$ in the following iteration.

%{\bf Quantifying ReLU Reduction in Each Steps} 
%\textcolor{red}{
%The amount of ReLU reduction in each steps varies across the networks and the spatial size of input images. Nevertheless, we found that ReLU culling reduces the ReLU count by a huge margin and hence a most effective step in DeepReDuce. For example, in ResNet18 this ranges from 2.42$\times$ (when $S_1$ is culled) to 8.4$\times$ (when $S_1$, $S_2$, and $S_4$ are culled). The other steps in DeepReDuce, systematically reduces the ReLU count by to reach a target ReLU budget.}

{\bf Complexity of DeepReDuce:} 
%In each stage-Culling iteration, ReLU Thinning and Reshaping are applied. When a new stage is Culled, we first apply Thinning to the remaining ReLU stages.After Thinning, we apply Reshaping to all stages usingchannel, fmap, and compound (i.e., both channel and fmap) scaling.
For each iteration Culling, Thinning, and Reshaping are applied individually,
resulting in five optimized networks per optimization iteration. 
Figure \ref{fig:BlockDiagramDeepReDuce} shows a single (initial) iteration of DeepReDuce.
DeepReDuce explores 5$\times(D-1)$ network architectures as we never Cull the most critical stage. Note that, irrespective of the depth of network, the number of stages varies between 3 to 5. 
In contrast, the NAS-based architecture search methods, including CryptoNAS, explore a large design space and train significantly more models. 
Thus, DeepReDuce is more efficient and effective than existing techniques.

\section{Methodology}
\label{sec:Methodology}

{\bf Network architecture:}
We apply DeepReDuce to standard ResNet18/34 architectures as defined in~\cite{he2016deep}, and also, on the non-residual networks VGGNet \cite{simonyan2014very} and MobileNets \cite{howard2017mobilenets}. 
%Since the datasets we use (CIFAR-100 and TinyImageNet) have smaller resolution images than ImageNet, we do not perform down sampling after the Conv1 layer or the first stage .

To show DeepReDuce networks outperform shallower ResNets we trained
ResNet10 and ResNet9 in Table~\ref{tab:ShallowNetComp}.
In these networks we removed half of the residual Blocks in each stage of ResNet18
(each stage now has only one residual Block).
Furthermore, for ResNet9, there is only one $3\times3$ convolution layer in the first residual Block of $S_1$.  
All other comparisons with state-of-the-art use reported results from respective papers.
 
\begin{table} [t] \centering
\caption{Optimizations applied (Culling, Thinning, and Reshaping) to Pareto points in Figure \ref{fig:ParetoFrontier}. Stages with ``$^\star$'' have only one Block inside the ReLU-stages. Acc. is top-1 accuracy for ResNet18 on CIFAR-100, and Lat. is inference time in seconds.}
%\fixme{Lines 8 and 10 are the same.
%it seems like we're just searching and not following the method.}}
\label{tab:ParetoPoints} 
\resizebox{0.5\textwidth}{!}{
\begin{tabular}{cccccccc}\toprule
\multirow{2}{*}{ Culled } & \multirow{2}{*}{ Thinning } & \multicolumn{2}{c}{ ReLU Reshaping } & \multirow{2}{*}{ \#ReLUs } & \multirow{2}{*}{ Acc.(\%) } & \multirow{2}{*}{ Lat.(s) } & \multirow{2}{*}{ Acc./ReLU }  \\
\cmidrule(lr{0.5em}){3-4}
& & Ch. & Fmap & & & & \\ \toprule 
\multicolumn{8}{c} {ResNet18 baseline model: \#ReLUs = 557.06K, top-1 accuracy (W/o KD) = 74.46\%} \\ \midrule
$S_1$ & NA & NA & NA & 229.38K & 76.22        &      4.61           & 0.332 \\
$S_1$+$S_4$ & NA & NA & NA & 196.61K & 75.51   &      3.94             & 0.384 \\
$S_1$ & $S_2$+$S_3$+$S_4$ & NA & NA & 114.69K & 74.72  & 2.38                & 0.651 \\
%$S_1$+$S_2$ & NA & NA & NA & 98.31K & 73.16        &     2.18          & 0.744 \\
$S_1$ & $S_2$+$S_3$+$S_4$ & 0.5$\times$ & NA & 57.34K & 72.68     & 1.37     & 1.27 \\ 
$S_1$+$S_4$  & $S_2$+$S_3$ & 0.5$\times$ & NA & 49.15K & 69.50          &   1.19       & 1.45 \\
$S_1$ & $S_2$+$S_3$+$S_4$ & NA & 0.5$\times$ & 28.67K & 68.68  &  0.74        & 2.40 \\
$S_1$+$S_4$ & $S_2$+$S_3$ & 0.5$\times$ & NA & 24.57K & 68.41  &   0.56      & 2.78 \\ 
$S_1$ & $S_2$+$S_3$+$S_4$ & 0.5$\times$ & 0.5$\times$ & 14.33K & 65.36& 0.52 & 4.56 \\
$S_1$+$S_4$ & $S_2$+$S_3$ & 0.5$\times$ & 0.5$\times$ & 12.28K & 64.97 & 0.45 & 5.29 \\
$S_1$ & $S_2^\star$+$S_3^\star$+$S_4^\star$ & 0.5$\times$ & 0.5$\times$ & 7.17K & 62.30 & 0.21 & 8.69 \\
\bottomrule
\end{tabular}}
\end{table}

{\bf Training process:}
Networks are trained using the following parameters:
an initial learning rate of $0.1$, 
mini-batch size of 128,
the momentum of $0.9$ (fixed), and 
$0.0004$ weight decay factor. 
We train networks for 120 epochs on both CIFAR-100 and TinyImageNet datasets. 
%\fixme{we train all tiny image net models for 120 epochs?}. 
The learning rate is reduced by a factor of $10$ every $30^{th}$ epoch. For training on CIFAR-10, we use cosine learning and train the networks for 150 epochs. 

When using knowledge distillation, we set the hyper-parameters,
temperature and relative weight to cross-entropy loss on hard targets 
as $4$ and $0.9$, respectively~\cite{hinton2015distilling,zagoruyko2016paying,cho2019efficacy}. 
For a fair comparison, we train all the networks with the same hyper-parameters 
and use the baseline model (without any ReLU dropping) as the teacher during KD.
For example, all the DeepReDuce-optimized ResNet18 networks and smaller ResNets,
such as ResNet10 and ResNet9, are trained with the baseline Full-ReLU ResNet18 as a teacher.

\begin{table} [t] \centering
\caption{Optimizations steps (Culling, Thinning, and Reshaping) for ReLU-optimized ResNet18 networks on TinyImageNet. Stages with ``$^\star$'' have only one Block inside the ReLU-stages. Acc. is top-1 accuracy, and Lat. is inference time in seconds.}
\label{tab:R18OnTinyImageNet} 
\resizebox{0.5\textwidth}{!}{
\begin{tabular}{cccccccc}\toprule
\multirow{2}{*}{ Culled } & \multirow{2}{*}{ Thinning } & \multicolumn{2}{c}{ ReLU Reshaping } & \multirow{2}{*}{ \#ReLUs } & \multirow{2}{*}{ Acc.(\%) } & \multirow{2}{*}{ Lat.(s) } & \multirow{2}{*}{ Acc./ReLU }  \\
\cmidrule(lr{0.5em}){3-4}
& & Ch. & Fmap & & & & \\ \toprule 
\multicolumn{8}{c} {ResNet18 baseline model: \#ReLUs = 2228.24K, top-1 accuracy (W/o KD) = 61.28\%} \\ \midrule
$S_1$ & NA & NA & NA & 917.52K & 64.66 & 17.16 & 0.070  \\
$S_1$ & $S_2$+$S_3$+$S_4$ & NA & NA & 458.76K & 62.26 & 8.87 & 0.136  \\
$S_1$+$S_2$ & NA & NA & NA & 393.24K & 61.65 & 7.77 & 0.157  \\
$S_1$ & $S_2$+$S_3$+$S_4$ & 0.5$\times$ & NA & 229.38K & 59.18 & 4.61 & 0.258  \\
$S_1$+$S_2$ & $S_3$+$S_4$ & NA & NA & 196.62K & 57.51 & 4.16 & 0.292  \\
$S_1$ & $S_2$+$S_3$+$S_4$ & NA & 0.5$\times$ & 114.69K & 56.18 & 2.47 & 0.490  \\
$S_1$+$S_2$ & $S_3$+$S_4$ & 0.5$\times$ & NA & 98.31K & 55.67 & 2.64 & 0.566  \\
$S_1$ & $S_2$+$S_3$+$S_4$ & 0.5$\times$ & 0.5$\times$ & 57.35K & 53.75 &  1.85 & 0.937 \\
$S_1$+$S_2$ & $S_3$+$S_4$ & NA & 0.5$\times$ & 49.16K & 49.00 & 1.325 & 0.997  \\
$S_1$ & $S_2^\star$+$S_3^\star$+$S_4^\star$ & 0.5$\times$ & 0.5$\times$ & 28.67K & 47.55 & 0.678 & 1.658  \\
$S_1$+$S_2$ & $S_3$+$S_4$ & 0.5$\times$ & 0.5$\times$ & 24.58K & 47.01 & 0.579 & 1.913  \\
$S_1$+$S_2$ & $S_3^\star$+$S_4^\star$ & 0.5$\times$ & 0.5$\times$ & 12.29K & 41.95 & 0.455 & 3.414  \\ 
\bottomrule
\end{tabular}}
\end{table}

{\bf Dataset:}
We perform our experiments on the CIFAR-100 \cite{2012KrizhevskyCIFAR} and TinyImageNet \cite{le2015tiny,yao2015tiny} datasets.  
CIFAR-100 has 100 output classes with 100 training and test images (resolution $32\times32$) per class.
TinyImageNet has 200 output classes with 500 training and 50 test/validation images (resolution $64\times64$) per class. 
We note that prior work on private inference~\cite{mishra2020delphi} has largely used smaller datasets like MNIST and CIFAR-10 in their evaluations, largely because the high costs of private inference make evaluations on large-scale images difficult.
%TinyImagenet is the most complex dataset used for private inference evaluations to date.

\textbf{Private inference protocol:}
We use the DELPHI~\cite{mishra2020delphi} protocol for private inference. 
DELPHI uses a secret sharing for linear layers and garbled circuits (GC) for ReLU layers. 
DELPHI optimizes the linear layer computations by moving cryptographic operations to an offline (preprocessing) phase. 
The protocol creates secret shares of the model weights during the offline phase (known before the client's input is available) and performs all linear operations over secret-shared data during the online phase.

\textbf{Threat model:}
We assume the same system setup and threat model as used by DELPHI~\cite{mishra2020delphi}, MiniOnn~\cite{liu2017oblivious}, and CryptoNAS~\cite{ghodsi2020cryptonas}.
This model assumes an honest-but-curious adversary.
We refer the interested reader to the referenced work for more details 
as we make no protocol changes and provide the exact same security guarantees.

\section{Results}  \label{sec:ExperimentalResults}

\subsection{DeepReDuce Pareto Analysis}
Figure~\ref{fig:ParetoFrontier} shows that 
DeepReDuce advances the ReLU count-accuracy Pareto frontier.
In this section, we present a detailed analysis and
quantify the benefit of networks along the frontier.
Table~\ref{tab:ParetoPoints} shows the details of the 
DeepReDuce Pareto points in Figure~\ref{fig:ParetoFrontier}.
Pareto points are shown in order of highest to lowest accuracy.
The primary takeaway from the table is that each of our optimizations (Culling, Thinning and Reshaping) are represented on the Pareto front, indicating that each is critical to obtaining the best results. 
Second, as mentioned in Section~\ref{sec:Reshaping}, Reshaping optimizations are most helpful at lower ReLU budgets, while Culling and Thinning dominate higher ReLU budget points.
This observation reaffirms that the order we apply our optimizations performs well.

We focus on CIFAR-100 first to best compare against prior work. Table~\ref{tab:R18OnTinyImageNet} provides the same data using the TinyImageNet dataset for ResNet18 and the criticality evaluation results are shown in Table \ref{tab:ResNetBlockDropout}. The data validates the effectiveness and hints at the general applicability of DeepReDuce.
Similar to the CIFAR-100 results, we found ReLUs in the initial layer 
(e.g., $S_1$) to be least critical while ReLUs in the intermediate layers (specifically ReLUs in penultimate stage) are most critical.

{\bf Accuracy per ReLU:}
Given that ReLU minimization and accuracy are competing objectives,
it is interesting to understand network design tradeoffs as accuracy per ReLU.
In the final column of Table~\ref{tab:ParetoPoints} and Table \ref{tab:R18OnTinyImageNet} 
we report each networks' accuracy per kilo-ReLU on CIFAR-100 and TinyImageNet datasets.
When ranking networks with respect to ReLU count (highest to lowest),
we observe accuracy per ReLU increases as ReLU count decreases.
In the extreme, the worst performing  CIFAR-100 network (Table~\ref{tab:ParetoPoints}) is 13.9\% less accurate than the most accurate network but also uses 32$\times$ fewer ReLUs, resulting in 26.2$\times$ more accuracy per kilo-ReLU. 
Similarly, the lowest accurate network on TinyImageNet (Table \ref{tab:R18OnTinyImageNet}) is 22.7\% less accurate but 74.7$\times$ fewer ReLU count than the most accurate network, resulting in 48.8$\times$ more accuracy per kilo-ReLU.

We believe this is a beneficial property for ReLU optimizations like DeepReDuce.
It implies that as fewer ReLUs are used, each contributes more to accuracy,
picking up the slack.
Moreover, while it takes many ReLUs to train a highly-accurate network,
accuracy degrades slowly as significant quantities of ReLUs are dropped.
%The gracefulness in decline is likely in part due to DeepReDuce's careful selection of
%which ReLUs to elide.
This suggests that a natural robustness to ReLU dropping may be a property of neural networks
that designers can leverage to optimize networks for private inference.

\begin{table} [t] \centering
\caption{Comparison of DeepReDuce and the state-of-the-art
in private inference: CryptoNAS \cite{ghodsi2020cryptonas} and DELPHI \cite{mishra2020delphi}.
Results show that DeepReDuce strictly outperforms both solutions at various ReLU counts.
Acc. is top-1 accuracy on CIFAR-100 and Lat. is inference time in seconds.}
\label{tab:SOTAcomparison} 
\resizebox{0.48\textwidth}{!}{
\begin{tabular}{cccccccccc} \toprule
\multirow{2}{*}{ } & \multicolumn{3}{c}{ SOTA } & \multicolumn{3}{c}{ DeepReDuce } & \multicolumn{3}{c}{ Improvement } \\
\cmidrule(lr{0.5em}){2-4}  
\cmidrule(lr{0.5em}){5-7} 
\cmidrule(lr{0.5em}){8-10} 
& ReLUs & Acc.(\%) & Lat.(s) & ReLUs & Acc.(\%) & Lat.(s) & ReLU  & Acc.(\%) & Lat.(s) \\ \toprule
\multirow{4}{*}{  \rotatebox[origin=c]{90}{CryptoNAS} } & 344K & 75.5 & 7.50 & 197K & 75.50 & 3.94 & 1.75$\times$ & 0.0 & 1.9$\times$ \\
& 100K & 68.7 & 2.30 & 28.6K & 68.70 & 0.738 & {\bf 3.5$\times$} & 0.0 & 3.1$\times$ \\
& 86K & 68.1 & 2.00 & 28.6K& 68.70 & 0.738 & 3$\times$ & 0.6 & 2.7$\times$ \\
& 50K & 63.6 & 1.67 & {\bf 12.3K} & {\bf 65.00} & {\bf 0.455} & 4$\times$ & 1.4 & 3.7$\times$ \\ \midrule
\multirow{3}{*}{ \rotatebox[origin=c]{90}{DELPHI} } & 300K & 68 & 6.5 & 28.7K & 68.70 & 0.738 & 10.5$\times$ & 0.7 & 8.8$\times$ \\
& 180K & 67 & 4.44 & 24.6K & 68.41 & 0.579 & 7.3$\times$ & 1.4 & 7.7$\times$ \\
& 50K & 66 & 1.23 & {\bf 49.2K} & {\bf 69.50} & {\bf 1.19} & 1$\times$ & {\bf 3.5} & 1$\times$ \\
\bottomrule
\end{tabular}}
\vspace{-2em}
\end{table}

%\begin{tabular}{c|cc|cc|cc} \toprule
%\multirow{2}{*}{ } & \multirow{2}{*}{ \#ReLUs } & \multirow{2}{*}{ Acc(\%) } & \multicolumn{2}{c|}{ DeepReDuce } & \multicolumn{2}{c}{ Improvement } \\
%& & & \#ReLUs & Acc (\%) & ReLU ($\downarrow$) & Acc(\%$\uparrow$) \\ \toprule
%\multirow{4}{*}{ CryptoNAS } & 344K & 75.5 & 196.61K & 75.50 & 1.75$\times$ & - \\
%& 100K & 68.7 & 28.67K & 68.70 & 3.5$\times$ & - \\
%& 86K & 68.1 & 28.67K & 68.70 & 3$\times$ & 0.60 \\
%& 50K & 63.6 & {\bf 12.29K} & {\bf 65.00} & {\bf 4$\times$} & 1.40 \\ \midrule
%\multirow{3}{*}{ DELPHI } & 300K & 68 & 28.67K & 68.70 & 10.5$\times$ & 0.70 \\
%& 180K & 67 & 24.58K & 68.41 & 7.3$\times$ & 1.41 \\
%& 50K & 66 & {\bf 49.15K} & {\bf 71.10} & 1$\times$ & {\bf 5.10} \\ 
\begin{figure}[t] \centering
\includegraphics[scale=0.45]{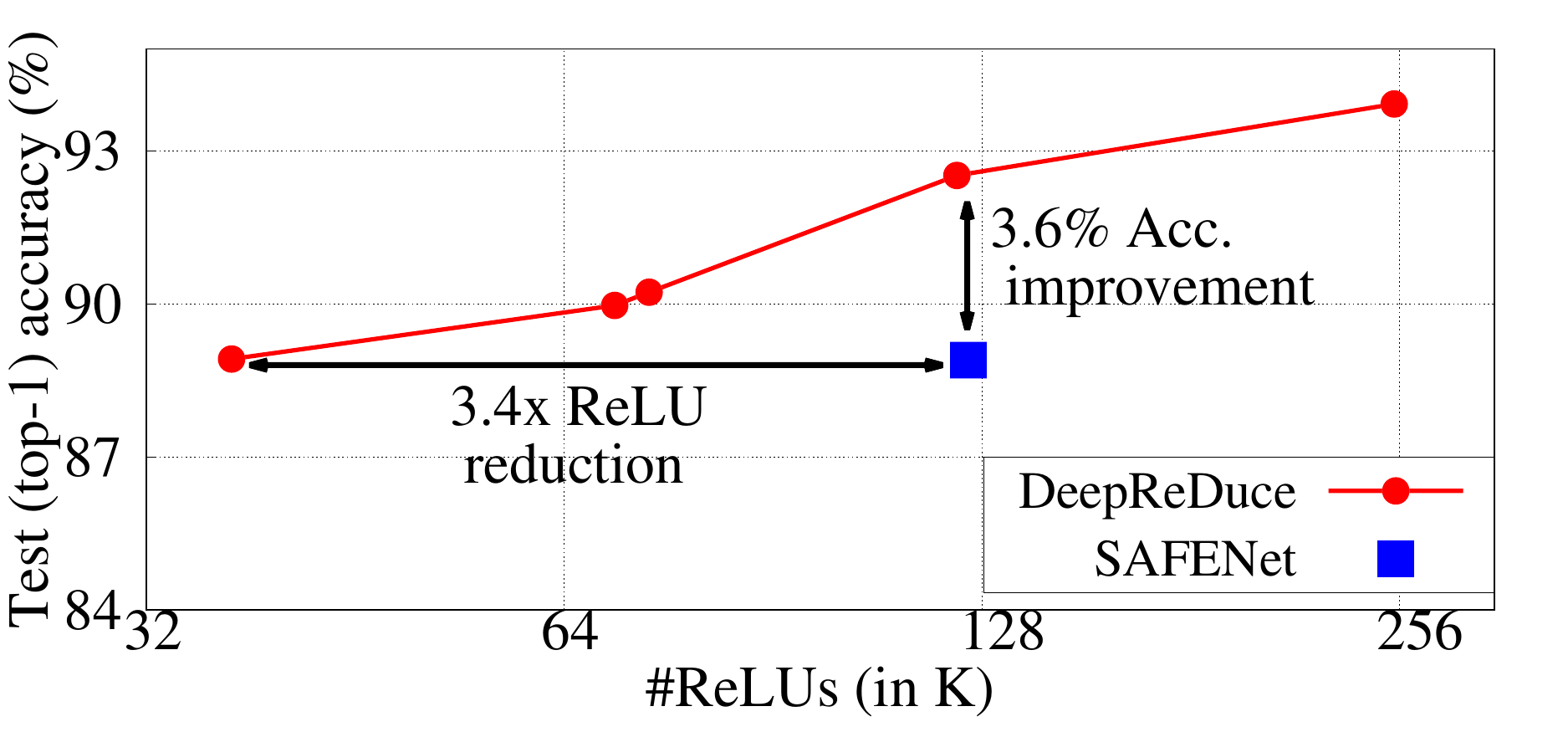}
\vspace{-2em}
\caption{
Performance comparison of VGG16 DeepReDuce model and SAFENet on CIFAR-10. The DeepReDuce-optimized VGG16 models outperform SAFENet-optimized VGG16 by a huge margin at both iso-accuracy and iso-ReLU.}
\vspace{-2em}
\label{fig:ParetoFrontierVGG16}
\end{figure}

\subsection{DeepReDuce Outperforms Prior Work}
Table~\ref{tab:SOTAcomparison} shows competing design points
for CryptoNAS and DELPHI.
We observe that CryptoNAS networks work well for high accuracy,
while DELPHI is best for small ReLU budgets.
To fairly compare, we only select DeepReDuce points that
offer benefit in both accuracy and ReLU count.

Starting with high-accuracy networks,
CryptoNAS needs 100K ReLUs to get an accuracy of 68.7\%;
DeepReDuce is able to match this accuracy using only 28.7K,
providing a savings of 3.5$\times$ ReLUs.
A DELPHI network needs 300K ReLUs to achieve similar accuracy,
which is 10.5$\times$ more ReLUs than DeepReDuce uses.
DELPHI is more competitive with networks achieving 67\% and 66\% accuracy.
In fact, DELPHI outperforms CryptoNAS when targeting a 50K ReLU budget,
reporting 66\% accuracy,
while CryptoNAS is only able to realize 63.6\% given the same ReLU budget.
With a target of 50K ReLUs, DeepReDuce optimizes a network that is 
(in absolute terms) 3.5\% more accurate than DELPHI.
Considering the Pareto set of merged DELPHI and CryptoNAS points,
DeepReDuce offers a maximum ReLU savings of 3.5$\times$ (iso-accuracy)
and a accuracy benefit of 3.5\% (iso-ReLU budget).
Finally, we note that CryptoNAS compares with and outperforms existing NAS methods for FLOP-optimized network design, and by extension DeepReDuce outperforms these methods as well.

We further compare DeepReDuce to SAFENet, a recently proposed \cite{lou2021safenet} fine-grained, channel-wise ReLU optimization targeting ReLU-heavy layers. 
SAFENet works by selectively substituting ReLUs with polynomials and uses different polynomials and approximation ratios across layers.
%ReLUs approximation ratio across the layers. 
Comparing the online latency for ResNet on CIFAR-100, we find DeepReDuce is {\em 12.86$\times$ faster} (0.56s vs 7.2s) and {\em 1.18\% more accurate} (68.68\% vs 67.5\%). For VGG16 on CIFAR-10, SAFENet reports 88.9\% accuracy with 56\% ReLUs approximated; even assuming all SAFENet approximated ReLUs are free, DeepReDuce provides a 3.4$\times$ ReLU reduction at iso-accuracy and 3.6\% more accuracy at iso-ReLU count (Figure \ref{fig:ParetoFrontierVGG16}). The criticality evaluation for VGG16 and optimization steps for the DeepReDuce Pareto points shown in Figure are \ref{fig:ParetoFrontierVGG16} are listed in Table \ref{tab:ReluCriticalityVGG16} and \ref{tab:ParetoPointsVGG16}, respectively, in Appendix \ref{SecAppendix:ParetoPointsVGG16}.

%We compare the performance of SAFENet-optimized ResNet (on CIFAR-100) and VGGNet (on CIFAR-10) with the respective DeepReDuce-optimized networks. 

\begin{table} [t] \centering
\caption{A comparison of DeepReDuce against smaller ResNet models on CIFAR-100. Results show that, iso-ReLU count and iso-accuracy, 
DeepReDuce consistently outperforms smaller ResNets.}
\label{tab:ShallowNetComp} 
\resizebox{0.49\textwidth}{!}{
\begin{tabular}{ccccc} 
 \toprule
 Network & \#Conv & \#ReLUs &  W/o KD(\%) & W/ KD(\%) \\ \toprule
ResNet10, $\alpha$=0.5 & 9 & 155.6K & 71.3 & 72.5\\
$S_2^{RT}$ + $S_3$ + $S_4^{RT}$ & 17 & 147.6K & 71.7 & 74.8 \\ \midrule
ResNet10, $\rho$=0.5 & 9 & 47.1K & 64.7 & 68.1\\
$S_2^{RT}$ + $S_3^{RT}$ & 11 & 49.2K & 67.8 & 71.0 \\ \midrule
ResNet9, $\alpha$=0.5, $\rho$=0.5 & 8 & 30.7K & 62.6 &	66.2 \\ 
$S_2^{RT}$ + $S_3^{RT}$ + $S_4^{RT}$, $\rho$=0.5 & {\bf 8} & {\bf 28.7K} & {\bf 64.4} & {\bf 68.5} \\
$S_2^{RT}$ + $S_3^{RT}$, $\alpha$=0.5 & 7 & 24.6K & 66.0 & 68.1 \\ \bottomrule
\end{tabular}} 
\vspace{-2em}
\end{table}

\subsection{DeepReDuce Outperforms Shallow ResNets} \label{subsec:ShallowResNet}

DeepReDuce's Culling and Thinning optimizations effectively reduce network depth. 
Thus, it is natural to ask why not simply train a smaller network?
We compared DeepReDuce against 
standard ResNet architectures with similar depth and ReLU counts. 
To compare fairly, we start with a baseline ResNet18 model and scale it down
to match the ReLU counts of DeepReDuce networks (see Section~\ref{sec:Methodology} for details).
To compare the performance at both iso-Layer and iso-ReLU, 
we merge the linear layers/Blocks in DeepReDuce networks at inference time. 
Since DeepReDuce uses KD, we also apply it to scaled ResNets for fair comparison. 
The results are shown in Table~\ref{tab:ShallowNetComp}. 

We first show that DeepReDuce outperforms ResNet10 models using both fmap and channel scaling.
When given the same number of layers (see bold row), DeepReDuce
uses two thousand fewer ReLUs and is 2.3\% more accurate than ResNet9.
DeepReDuce uses fewer convolution layers when dropping a ReLU connecting two
convolutions layers, which allows them to be combined or merged.
Moreover, we can continue to scale down DeepReDuce to use fewer ReLUs (24.6K)
and still be 1.9\% more accurate than the smaller ResNet9 model.
Thus, we conclude that DeepReDuce outperforms simply training smaller networks.

\subsection{DeepReDuce Outperforms Channel Pruning}

While DeepReDuce and channel pruning have different optimization objectives, channel pruning also reduces ReLUs. 
We compare DeepReDuce  with a recent state-of-art channel pruning method \cite{he2020learning} and compare the two methods in terms  ReLUs, FLOPs, and accuracy. 
Since authors in aforementioned paper report results using ResNet56 as the baseline, we ran DeepReDuce on the same network. 
(Table \ref{tab:ReluCriticalityR56} in Appendix \ref{SecAppendix:ReluCriticalityInR56} shows the per-stage criticality for ResNet56; 
stage $S1$ ($S3$) is the least (most) critical stage for this network.)

Table \ref{tab:PruningSOTAComp} compares DeepReDuce with channel pruning on the CIFAR-10 and CIFAR-100 datasets.  DeepReDuce has the higher accuracy (0.73\% and 2.83\% more accurate on CIFAR-10 and CIFAR-100, respectively) and uses $1.4\times$ fewer ReLUs compared to channel pruning. At a slightly lower accuracy (0.18\% less) on CIFAR-10, 
DeepReDuce's reduction in ReLUs over channel pruning increases to more than $2\times$ (147.5K vs. 311.7K).

DeepReDuce performs even better on CIFAR-100. 
It is 1\% more accurate (71.68\% vs. 70.83\%) with 2$\times$ fewer ReLUs compared to channel pruning. 
In all our comparisons we observed that DeepReDuce has $~1.4\times$ more FLOPs compared to channel pruning; 
this is not a problem for DeepReDuce because FLOPs are effectively free for private inference run-time as noted by~\cite{ghodsi2020cryptonas}.
We conclude that FLOP-count oriented network optimizations are very different from ReLU-oriented network optimizations.

\begin{table} [t] \centering
\caption{Performance comparison of channel pruning \cite{he2020learning} and DeepReDuce for FLOPs and ReLU saving, and accuracy drop on ResNet56 with CIFAR-10 (C10) and CIFAR-100 (C100) datasets. DeepReDuce models save significantly higher \#ReLUs at similar FLOPs saving and accuracy drop.}
\label{tab:PruningSOTAComp} 
\resizebox{0.49\textwidth}{!}{
\begin{tabular}{ccp{1.2cm}p{1cm}p{0.8cm}cc}\toprule
& Method & Baseline Acc.(\%) & Pruned Acc.(\%) & Acc. $\downarrow$(\%) & FLOPs & ReLUs \\ \toprule
\multirow{3}{*}{ \rotatebox[origin=c]{90}{C10} } & Ch. pruning  & 93.59 & 93.34 & -0.25 & 59.1M & 311.7K \\ \cline{2-7}
& \multirow{2}{*}{DeepReDuce} & \multirow{2}{*}{93.48} & 94.07 & +0.59 & 87.7M & 221.2K \\
& & & 93.16 & -0.32 & 66.5M & 147.5K  \\ \midrule
\multirow{3}{*}{ \rotatebox[origin=c]{90}{C100} } & Ch. pruning  & 71.41 & 70.83 & -0.58 & 60.8M & 311.7K  \\ \cline{2-7}
& \multirow{2}{*}{DeepReDuce} & \multirow{2}{*}{70.93} & 73.66 & +2.57 & 87.7M & 221.2K \\
& & & 71.68 & +0.59 & 66.5M & 147.5K  \\
\bottomrule
\end{tabular}}
\vspace{-2em}
\end{table}

\subsection{DeepReDuce Network Inference Latencies}

We conclude by showing the inference speedup time improvements offered by DeepReDuce.
Experiments were run to measure the latency for DeepReDuce optimized models using the same experimental setup and private inference protocol as DELPHI ~\cite{mishra2020delphi}.
Table~\ref{tab:ParetoPoints} and Table~\ref{tab:R18OnTinyImageNet} present latency results of ResNet18 on CIFAR-100 and TinyImageNet, respectively.  
As expected, we observe that inference latency strongly correlates with a network's ReLU count. 
For example, at iso-accuracy (on CIFAR-100), we achieve a 3.7$\times$ latency reduction compared to CryptoNAS. 
Compared to DELPHI, and assuming iso-latency, DeepReDuce offers 3.5\% accuracy improvement on CIFAR-100. 
The fastest DeepReDuce model on CIFAR-100, i.e., the one with the fewest ReLUs, takes 455mS per inference with an accuracy of 65\%.  
Similarly, on TinyImageNet, we achieve 59.18\% top-1 accuracy with a latency of 4.6S.
While advancing the state-of-the-art, these inference times are still too high to meet real-time requirements 30-60 FPS~\cite{edgeFB}, and more work is needed to achieve the ultimate goal of real-time private inference.

\subsection{Generality Case Study: MobileNets}
\begin{figure}[t] \centering
\includegraphics[scale=0.45]{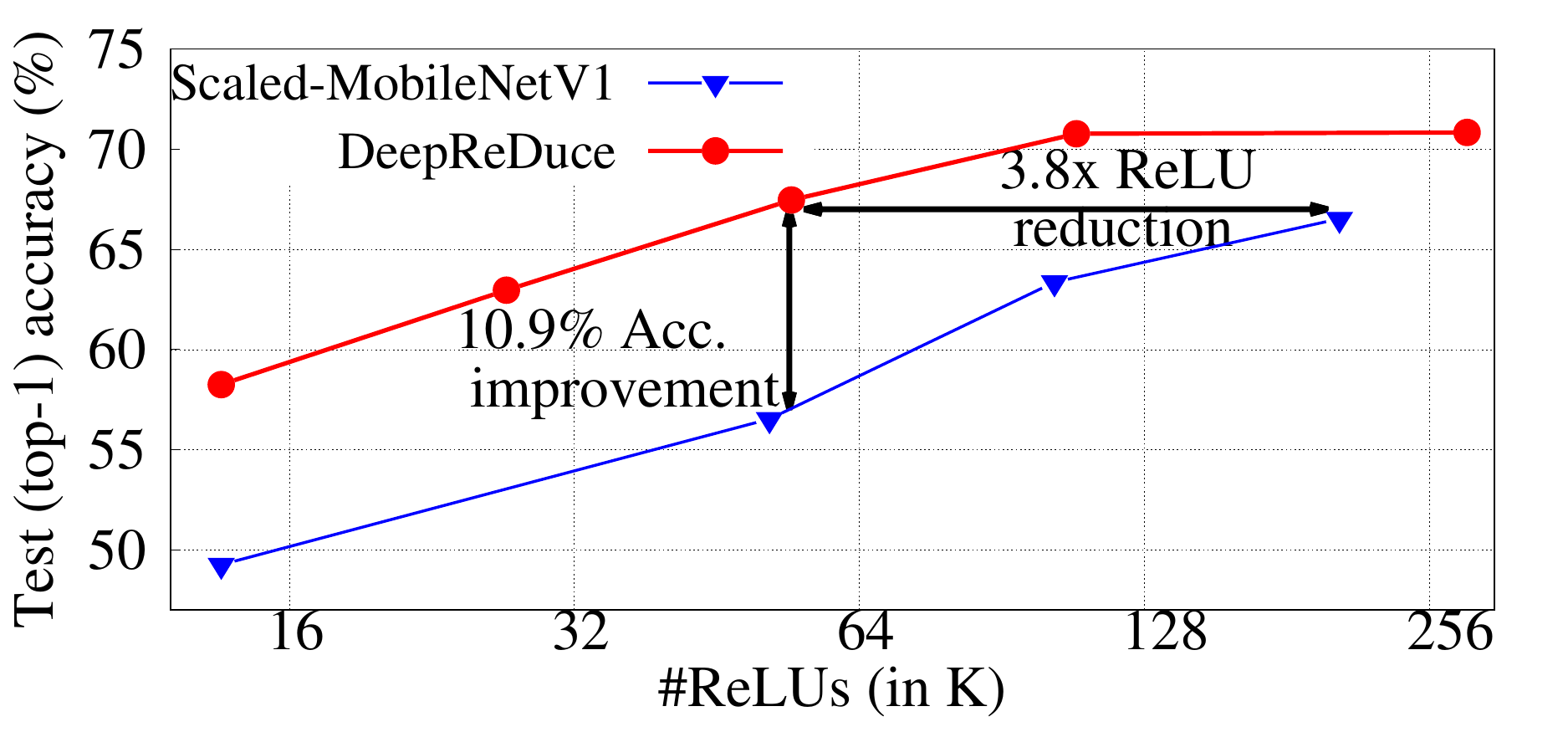}
%\vspace{-2.5em}
\caption{
Performance comparison of DeepReDuce models and (channel/fmap-resolution) scaled MobileNetV1 on CIFAR-100. DeepReDuce optimized models outperform the scaled MobileNetV1 by a huge margin at both iso-accuracy and iso-ReLU.}
%\vspace{-2em}
\label{fig:ParetoFrontierMV1}
\end{figure}

Here we examine the generality of DeepReDuce using MobileNetV1 \cite{howard2017mobilenets}.
We chose to study MobileNet as it is very different from ResNet---the convolution layers
do not use residuals and the Depthwise architecture is FLOP optimized, which we believe is a poor match for the ReLU costs of private inference.
%types (a non-residual network with FLOPs-optimized Depthwise separable convolution) are significantly different than that of ResNet models.
We first evaluate the ReLUs' criticality (see Table \ref{tab:CriticalityInMobileNets} in Appendix \ref{SecAppendix:ParetoPointsMV1}) and then compare the performance of ReLU-optimized DeepReDuce models with conventionally (channel/fmap-resolution) scaled MobileNetV1 models. For fair comparison, we use KD for scaled-MobileNetV1 models where teacher is Full-ReLU baseline MobileNetV1 and hence, all the results reported in Figure\ref{fig:ParetoFrontierMV1} are with KD.

Results are shown in Figure \ref{fig:ParetoFrontierMV1} and the optimization steps for all DeepReDuce models are listed in Table \ref{tab:ParetoPointsMV1} in Appendix \ref{SecAppendix:ParetoPointsMV1}. 
The substantial gain, 10.9\% improvement in accuracy at iso-ReLU and 3.8$\times$ ReLU reduction at iso-accuracy (see Figure \ref{fig:ParetoFrontierMV1}) shows the effectiveness of DeepReDuce ReLU optimization on MobileNetV1.

Thus, while residual connections benefit DeepReDuce, they are not
a necessity as DeepReDuce also performs well when residual connections are eliminated from the network and the non-residual networks such as MobileNets and VGG16 (Figure \ref{fig:ParetoFrontierVGG16}).

% \subsection{ResNet with KD Outperforms State-of-the-art in Private Inference}

% We selected ResNet (as baseline) for our experimentation because of its efficacy in various computer-vision tasks, which is also corroborated by the results shown in  Figure \ref{fig:ParetoFrontierWithKD}. The (scaled) ResNets trained with KD, under the supervision of full ReLU ResNet18 as a teacher, surpass the accuracy of CryptoNAS and DELPHI at iso-ReLU counts. This suggest that with appropriate depth, width, and fmaps' resolution scaling in conjunction with the KD, the classical ResNets can outperform the sophisticated NAS-based ReLU-optimization methods such as CryptoNAS and DELPHI. 

% However, as illustrated in Figure \ref{fig:ParetoFrontierWithKD}, with lower ReLU counts the accuracy gain through KD in ResNets start diminishing. In contrast,  DeepReDuce models maintain the ReLU-accuracy trade-off even at very low ReLU counts and their performance does not degrade sharply.  

%\input{19Figure_ParetoFrontierWithKD}

%\input{06Generality}
\section{Related Work}
\label{sec:relatedwork}

%\subsection{Private Inference}
CryptoNets~\cite{gilad2016cryptonets} was the first to demonstrate using homomorphic encryption to protect client data during inference. 
%CryptoNets uses polynomial activations to polynomials (quadratic function). 
SecureML~\cite{mohassel2017secureml} focuses on privacy-preserving training of several machine learning models, including neural networks using a two-server protocol. 
While \cite{mohassel2017secureml} supports inference as well, it incurs high overheads by relying on generic MPC protocols. MiniONN~\cite{liu2017oblivious} generates multiplication triplets for each multiplication in a linear layer and combines that with GC protocol for ReLU activation functions. 
Gazelle~\cite{juvekar2018gazelle} uses an optimized HE scheme for linear layers and GC for non-linear layers. DELPHI~\cite{mishra2020delphi} further optimizes this protocol by moving the heavy cryptographic operations to an offline preprocessing phase and using only secret sharing for linear layers online. 
In DELPHI, select ReLU layers are replaced with quadratic functions and the authors propose a neural architecture search method (NAS) to determine which ReLU layers to replace. 
CryptoNAS~\cite{ghodsi2020cryptonas} defines a ReLU budget for private inference task and aims to find the best networks for a budget using NAS. A recent work SAFENet \cite{lou2021safenet} selectively replaces the channel-wise ReLUs with multiple degree polynomials and uses layer-wise mixed precision. 

% \subsection{Effectiveness of ReLU} 
% Glorot et al.~\cite{glorot2011deep} first proposed the rectifier linear activation function, and the efficacy of ReLU for better convergence of deeper neural networks came into the limelight with the advent of AlexNet~\cite{krizhevsky2012imagenet}. Further, different versions of ReLU have been proposed mainly to overcome the issue of dead neurons and/or gradient vanishing/exploding \cite{maas2013rectifier,he2015delving,clevert2015fast,xu2015empirical,shang2016understanding,chen2020dynamic}.
% However, Zhao et al.~\cite{zhao2018reThinking} claimed that ReLU activations after each convolution layers creates redundancy and could lead to poor generalization. They showed a noticeable gain in accuracy when the network is trained with ReLUs in alternate convolution layers, which lines up with our empirical observation for dropping the ReLUs from alternate convolution layers in ReLU-stages.

% \subsection{Channel Pruning}
% Channel pruning is an effective way to reduce the FLOPs/parameter counts, and unlike the uniform channel scaling \cite{howard2017mobilenets,tan2019efficientnet}, ~\cite{wang2020apq,liu2019metapruning,he2018amc} proposed automatic pruning; however, their effectiveness for ReLU saving is yet to be examined since the distribution of FLOPs/parameters are substantially different from that of the ReLUs in the networks.

%\input{16Figure_StageWiseReluPercentage}

\section{Conclusion and Future Work} \label{sec:Conclusion}
This paper develops the concept of ReLU dropping  
as an effective method for tailoring networks for private inference.
The DeepReDuce method carefully prioritizes and
removes ReLUs based on how critical they are, and users are given a
wide range of networks that trade ReLU count and accuracy.
Evaluating DeepReDuce on CIFAR-100 shows 3.5\% improvement at iso-ReLU and 3.5$\times$ ReLU reduction at iso-accuracy. 
We also perform the experiments on TinyImageNet, which is uncommon for private inference work given the long runtimes. 
While we advance the state-of-the-art, we still fall short of the ultimate goal of real-time private inference.
We expect DeepReduce to inspire future optimizations and new ways of Thinking about network design to achieve this goal.

\section*{Acknowledgements}
This work was supported in part by the Applications Driving Architectures (ADA) Research Center, a JUMP Center co-sponsored by SRC and DARPA.
This research was also developed with funding from the Defense Advanced Research Projects Agency (DARPA),under the Data Protection in Virtual Environments (DPRIVE) program, contract HR0011-21-9-0003.
The views, opinions and/or findings expressed are those of the author and should not be interpreted as representing the official views or policies of the Department of Defense or the U.S. Government.

%\section*{Acknowledgements}

\nocite{langley00}

\bibliography{MyRef}
\bibliographystyle{icml2021}

\newpage
\appendix
\clearpage

%\section{Computation and Communication Overheads in ReLU Computation} \label{sec:ReluOverhead}
%
%The computation and communication (between server and client in two party computation settings) overheads of individual  ReLU and convolution (Conv) operation is shown in Table \ref{tab:ReluOverheads}. These (amortized) values are calculated according to the Table 1 and Table 3 in  DELPHI \cite{mishra2020delphi}. Clearly, ReLU operation has orders of magnitude higher computation and communication overheads (especially in online phase) compared to the convolution operations. Therefore, removing ReLUs from networks (with minimal impact on accuracy)  is a paramount importance for the private inference. 
%\input{42AppendixTable_ReluOverheadComp}

\section{Channel Scaling Vs. Feature map Scaling} \label{secAppendix:ChannelVsFmapScaling}

For extremely lower ReLU budgets, we use a combination of channel scaling and fmaps' resolution scaling. Since reducing  fmaps' resolution (each spatial dimensions of fmaps) by 2$\times$ ($\rho$=0.5) decreases the ReLU count by 4$\times$, we first use channel scaling ($\alpha$ = 0.5) for reducing the ReLU count by 2$\times$. Further, for 4$\times$ reduction in ReLU count, we prefer to use fmaps' resolution scaling ($\rho$=0.5) over channel scaling ($\alpha$=0.25) since the former results in more accurate networks, as illustrated in Table \ref{tab:AlphaVsRhoComp}. Unlike fmap resolution scaling, channel scaling reduces the parameter count along with the ReLU count, which may reduce the expressive power of a network. Hence, the network is more accurate with fmaps' resolution scaling.  
  
\begin{table} [htbp]
\caption{Performance comparison for ReLU optimization using channel scaling ($\alpha<$1) and fmap-resolution scaling ($\rho<$1). Baseline models have $\alpha$=1 and $\rho$=1. At iso-ReLU, accuracy (w/ KD) of the the fmap-resolution scaled models is higher than the channel scaled models.}
\label{tab:AlphaVsRhoComp}
\centering 
\resizebox{0.49\textwidth}{!}{
\begin{tabular}{ccccc} 
 \toprule
\multirow{2}{*}{Network} & \multirow{2}{*}{\#Conv} & \multirow{2}{*}{\#ReLUs} & \multicolumn{2}{c}{CIFAR-100} \\
\cmidrule(lr{0.5em}){4-5}  
& & & {W/o KD (\%)} & {W/ KD (\%)} \\
 \toprule
ResNet18 (baseline) & 17 & 557K & 74.46 & 76.94 \\
ResNet18; $\alpha$=0.25, $\rho$=1  & 17 & 139K & 68.17 & 70.19 \\
ResNet18; $\alpha$=1, $\rho$=0.5 & 17 & 139K & 68.47 & 72.72 \\ \midrule
ResNet10 (baseline)& 9 & 311K & 74.10 & 76.69 \\
ResNet10; $\alpha$=0.25, $\rho$=1 & 9 & 78K & 66.69 & 66.88 \\ 
ResNet10; $\alpha$=1, $\rho$=0.5 & 9 & 78K & 66.67 & 71.86 \\ \midrule
ResNet6 (baseline) & 5 & 188K & 68.86 & 69.58 \\
ResNet6; $\alpha$=0.25, $\rho$=1 & 5 & 47K & 57.64 & 56.9 \\
ResNet6; $\alpha$=1, $\rho$=0.5 & 5 & 47K & 64.74 & 68.09 \\ 
\bottomrule
\end{tabular}} 
\end{table}

\section{VGGNet DeepReDuce Pareto Points} \label{SecAppendix:ParetoPointsVGG16}

We do not remove ReLUs from fully connected (FC) layers as FC account only 8.192K ReLUs and training networks without FC ReLUs is challenging. The results are shown in Table \ref{tab:ReluCriticalityVGG16}. Unlike ResNets and MobileNets, ReLUs in $S_5$ are least critical and that of the $S_1$ is moderate. 

\begin{table} [htbp]
\caption{Stage-wise ReLUs' criticality  in VGG16 evaluated on CIFAR-10. $S_5$ is least critical while $S_2$ and $S_3$ are most critical.}
\label{tab:ReluCriticalityVGG16}
\centering 
\resizebox{0.49\textwidth}{!}{
\begin{tabular}{ccccc} \toprule
Net & \#ReLUs & W/o KD (\%) & W/ KD (\%) & $C_k$ \\ \toprule
%Baseline & 284.7 & 93.71 & & \\ \midrule
$S_1$ + FC-ReLUs & 139.3K & 82.0 & 81.4 & 10.90 \\
$S_2$ + FC-ReLUs & 73.7K & 86.1 & 85.3 & 14.31 \\
$S_3$ + FC-ReLUs & 57.3K & 86.4 & 85.1 & 14.40 \\
$S_4$ + FC-ReLUs & 32.8K & 77.1 & 77.7 & 9.17 \\
$S_5$ + FC-ReLUs & 14.3K & 63.9 & 66.0 & 0.00 \\ 
\bottomrule
\end{tabular} }
\end{table}

The ReLU optimizations step for the Pareto points in Figure \ref{fig:ParetoFrontierVGG16} are listed in Table \ref{tab:ParetoPointsVGG16}.  Models are the ReLU-optimized versions (Thinned and Reshaped) of two Culled networks: (1) stages $S_4$ and $S_5$ are Culled and (2) stages $S_1$, $S_4$ and $S_5$ are Culled. 

\begin{table} [htbp]
\caption{Optimization steps for MobileNetV1 DeepReDuce models shown in Figure  \ref{fig:ParetoFrontierVGG16}}
\label{tab:ParetoPointsVGG16}
\centering 
\resizebox{0.49\textwidth}{!}{
\begin{tabular}{lcc} \toprule
Optimization Steps & \#ReLUs & Acc.(\%) \\ \toprule
$S_1$ + $S_2$ + $S_3$ + FC & 253.95K & 93.92 \\
$S_2$ + $S_3$ + FC & 122.88K & 92.52 \\
$S_2^{RT}$ + $S_3^{RT}$ + FC & 73.73K & 90.23 \\
$S_1^{RT}$ + $S_2^{RT}$ + $S_3^{RT}$ + FC, $\alpha$=0.5 & 69.63K & 89.97 \\
$S_2^{RT}$ + $S_3^{RT}$ + FC, $\alpha$=0.5 & 36.86K & 88.92 \\   \bottomrule
\end{tabular} }
\end{table}

\section{ReLUs' Criticality and Pareto Points for MobileNets} \label{SecAppendix:ParetoPointsMV1}

We evaluate the ReLUs' criticality in MobileNetV1 \cite{howard2017mobilenets} and MobileNetV2 \cite{sandler2018mobilenetv2}) on the CIFAR-100. The results are shown in Table \ref{tab:CriticalityInMobileNets}.  We observed the similar trend as ResNet18 and ResNet34 on CIFAR-100/TinyImageNet (shown in Tables \ref{tab:ReluHeteroR18} and \ref{tab:ResNetBlockDropout}), accuracy differs significantly across stages and $S_1$ ($S_4$) ReLUs are least (most) critical.
\begin{table} [htbp] \centering
\caption{Stage-wise criticality of ReLUs in MobileNetV1 and MobileNetV2 evaluated on CIFAR-100. FR is baseline with Full-ReLU ($S_1$+$S_2$+$S_3$+$S_4$+$S_5$). ReLUs in $S_1$ ($S_4$) are least (most) critical.}
\label{tab:CriticalityInMobileNets} 
\resizebox{.5\textwidth}{!}{
\begin{tabular}{ccccccccc} \toprule
\multirow{2}{*}{ Net } & \multicolumn{4}{c}{ MobileNetV1 } & \multicolumn{4}{c}{ MobileNetV2 } \\ 
\cmidrule(lr{0.5em}){2-5}  
\cmidrule(lr{0.5em}){6-9} 
& \#ReLUs & W/o KD(\%) & W/ KD(\%) & $C_k$ &\#ReLUs & W/o KD(\%) & W/ KD(\%) & $C_k$ \\ \toprule
FR & 411.6K & 67.58 & - &    -   &425.6K & 68.46 & -     &  - \\ 
$S_1$ & 131.1K & 33.06 & 34.16 & 0.00 & 196.6K & 37.82 & 34.25 & 0.00 \\
$S_2$ & 114.7K & 49.64 & 50.65 & 11.83 & 110.6K & 49.83 & 46.93 & 9.12 \\
$S_3$ & 57.3K & 55.56 & 54.20 & 15.09 & 58.4K & 54.74 & 53.06 & 14.15 \\
$S_4$ & 94.2K & 57.37 & 61.10 & 19.60 & 27.6K & 57.08 & 57.28 & 18.26 \\
$S_5$ & 14.3K & 42.32 & 45.45 & 9.37 & 32.4K & 48.42 & 50.49 & 12.73 \\  \bottomrule
\end{tabular} }
\end{table}

The Pareto points of DeepReDuce models for MobileNetV1 (CIFAR-100) are shown in Figure \ref{fig:ParetoFrontierMV1}. The optimization steps for all DeepReDuce models are list in the Table \ref{tab:ParetoPointsMV1}.

First, in step 1 of DeepReDuce (Figure \ref{fig:BlockDiagramDeepReDuce}), we Culled the least critical stage $S_1$. In step 2 of ReLU Thinning, we had two ways to remove the ReLUs from alternate layers, either from $3\times3$ depthwise convolution layer or $1\times1$ pointwise convolution layer. When downsampling is performed in $3\times3$ depthwise convolution layer, the ReLU count of both the layers are not equal. More precisely, \#ReLUs in the $1\times1$ pointwise convolution is twice as that in the preceding $3\times3$ depthwise conv.

We empirically found that removing ReLUs from $3\times3$ depthwise conv layer yields more accurate iso-ReLU models. We suspect this is because $3\times 3$ depthwise convolutions perform filtering (feature learning) and $1\times1$ pointwise convolutions perform feature aggregation \cite{howard2017mobilenets}, the ReLUs in the former layer is more critical for accuracy.

\begin{table} [htbp]
\caption{Optimization steps for MobileNetV1 DeepReDuce models shown in Figure  \ref{fig:ParetoFrontierMV1}}
\label{tab:ParetoPointsMV1}
\centering 
\resizebox{0.49\textwidth}{!}{
\begin{tabular}{lcc} \toprule
Optimization Steps & \#ReLUs & Acc.(\%) \\ \toprule
$S_2$ + $S_3$ + $S_4$ + $S_5$ & 280.60K & 70.83 \\
$S_2^{RT}$ + $S_3^{RT}$ + $S_4^{RT}$ + $S_5^{RT}$ & 108.54K & 70.77 \\
$S_2^{RT}$ + $S_3^{RT}$ + $S_4^{RT}$ + $S_5^{RT}$, $\alpha$=0.5 & 54.27K & 67.46 \\
$S_2^{RT}$ + $S_3^{RT}$ + $S_4^{RT}$ + $S_5^{RT}$, $\rho$=0.5 & 27.14K & 62.96 \\
$S_2^{RT}$ + $S_3^{RT}$ + $S_4^{RT}$ + $S_5^{RT}$, $\alpha$=0.5, $\rho$=0.5 & 13.57K & 58.25 \\  \bottomrule
\end{tabular} }
\end{table}

\section{ReLU Criticality in ResNet56} \label{SecAppendix:ReluCriticalityInR56}

We examine the stage-wise criticality of ReLUs in ResNet56 and results are shown in Table \ref{tab:ReluCriticalityR56}.

\begin{table} [htbp]
\caption{Stage-wise criticality of ReLUs in ResNet56 evaluated on CIFAR-100. $S3$ is most critical and $S1$ is least critical.}
\label{tab:ReluCriticalityR56}
\centering 
%\resizebox{0.49\textwidth}{!}{
\begin{tabular}{ccccc} \toprule
Stages & \#ReLUs & W/o KD (\%) & W/ KD (\%) & $C_k$ \\ \toprule
S1 & 311.3K & 57.92 & 59.45 & 0.0 \\
S2 & 147.5K & 65.62 & 67.97 & 6.0 \\
S3 & 73.73K & 65.36 & 69.22 & 7.2 \\ \bottomrule
\end{tabular} 
\end{table}

\section{Layer-wise Distribution of ReLUs} \label{secAppendix:LayerWiseOpsInDNNs}

\begin{figure*}
\begin{subfigure}{1.0\textwidth}
\includegraphics[scale=0.27]{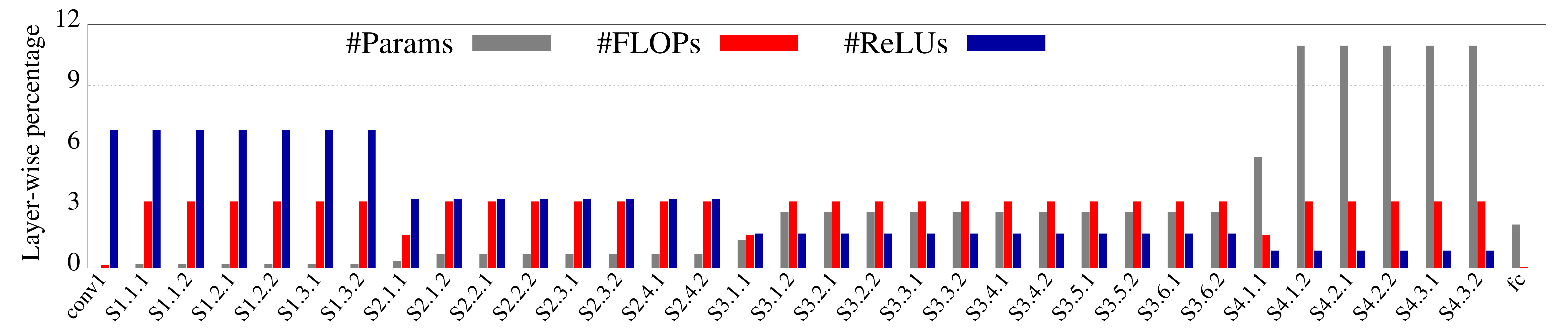}
\caption{Layer-wise distribution of parameter, FLOPs, and ReLUs in ResNet34 \cite{he2016deep}.} \label{fig:LayerWiseReluInOtherDNNs_a}
\end{subfigure} \\
\begin{subfigure}{1.0\textwidth}
\includegraphics[scale=0.27]{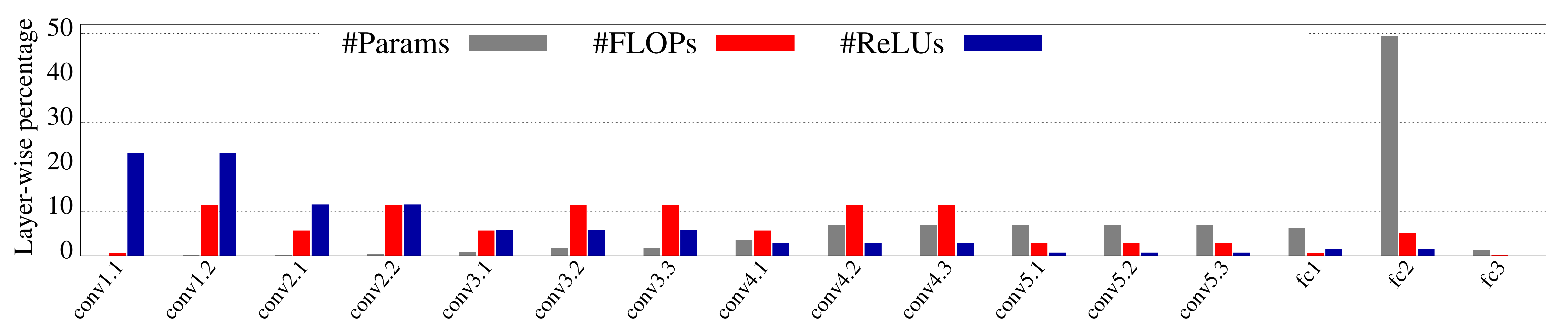}
\caption{Layer-wise distribution of parameter, FLOPs, and ReLUs in VGG16 \cite{simonyan2014very}.} \label{fig:LayerWiseReluInOtherDNNs_b}
\end{subfigure} \\
\begin{subfigure}{1.0\textwidth}
\includegraphics[scale=0.27]{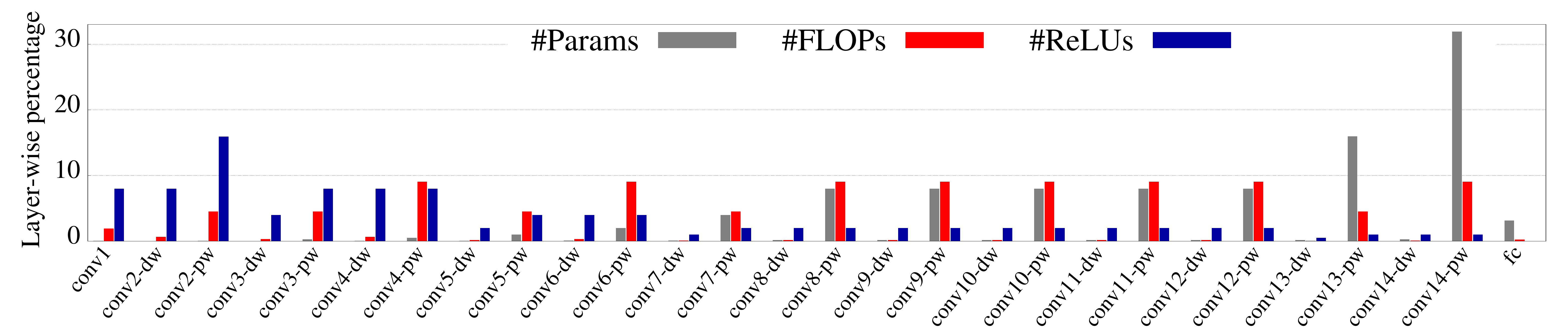}
\caption{Layer-wise distribution of parameter, FLOPs, and ReLUs in MobileNetV1 \cite{howard2017mobilenets}.} \label{fig:LayerWiseReluInOtherDNNs_c}
\end{subfigure} \\
\caption{Layer-wise percentage of parameter, FLOPs, and ReLUs in various DNNs. FLOPs are evenly distributed, parameters (ReLUs) are increases (decreases) from initial to deeper layers.} 
\label{fig:LayerWiseReluInOtherDNNs}
\end{figure*}

We show the layer-wise distribution of FLOPs, parameters, and ReLU count in the standard networks such as ResNet34 \cite{he2016deep}, VGG16 \cite{simonyan2014very}, MobileNetV1 \cite{howard2017mobilenets}, and MobileNetV2 \cite{sandler2018mobilenetv2} in Figure \ref{fig:LayerWiseReluInOtherDNNs}. Consistent with ResNet18 (Figure \ref{fig:LayerWiseReluInDNNs}), the FLOPs are evenly distributed, more parameters are used in deeper layers, and ReLUs are mostly in initial layers of the networks. Thus, the ReLU reduction in initial layers has a significantly greater impact on the overall ReLU count of these networks. Moreover, the stark difference between the ReLU distribution and FLOPs/parameter distribution indicates that ReLU optimization cannot be ensured through the popular FLOPs/parameters pruning techniques. 

%\begin{figure*}[htbp] 
%\includegraphics[scale=0.27]{Figures/LayerWiseOps_R34} 
%%\includegraphics[scale=0.27]{Figures/LayerWiseOps_Vgg16}   \\ 
%%\includegraphics[scale=0.16]{Figures/LayerWiseOps_MV1} 
%%\includegraphics[scale=0.18]{Figures/LayerWiseOps_MV2}
%\caption{
%Layer-wise distribution of parameter, FLOPs, and ReLUs for ResNet34 (top-left), and VGG16 (top-right); and MobileNetV1 (bottom left) and MobileNetV2 (bottom right). FLOPs are evenly distributed, parameters (ReLUs) are increases (decreases) from initial to deeper layers. }
%\label{fig:LayerWiseReluInOtherDNNs}
%\end{figure*}

%\newpage
%\input{11Rebuttal}

\end{document}